%% file: main.tex
\documentclass{article}

\PassOptionsToPackage{numbers,compress}{natbib}
\usepackage[main,final]{neurips_2026}

\input{preamble}
\input{defs}

\title{Latent Riemannian Flow Matching for Geometry-Grounded 3D Foundation Models}

\author{%
  Lisa Weijler \\
  TU Wien \\
  \And
  Irene Ballester \\
  TU Wien \\
  \And
  Guofeng Mei \\
  Fondazione Bruno Kessler \\
  \AND
  Tolga Birdal \\
  Imperial College London \\
  \And
  Pedro Hermosilla \\
  TU Wien \\
}

\begin{document}

\maketitle

\input{sec/0_abstract}
\input{sec/1_intro}

\input{sec/2_related_work}
\input{sec/3_methods}

\input{sec/4_experiments}
\input{sec/5_conclusion}

\bibliographystyle{plainnat}
\bibliography{main}

\input{sec/X_supp}

\newpage

\end{document}

%% file: preamble.tex
\usepackage{graphicx}
\usepackage{booktabs}
\usepackage{acronym}
\usepackage{algorithm}
\usepackage{algpseudocode}
\usepackage{wrapfig}
\usepackage{makecell}
\usepackage{amsmath}
\usepackage{mathtools}
\usepackage{caption}
\usepackage{tabularray}
\usepackage{multirow}
\usepackage{comment}
\usepackage{bm}
\usepackage{colortbl}
\usepackage[dvipsnames,table]{xcolor}
\usepackage{pifont}
\usepackage{adjustbox}  %
\usepackage{subcaption}
\usepackage{tikz}
\usetikzlibrary{arrows.meta}

\usepackage[accsupp]{axessibility}  %

\usepackage{hyperref}
\usepackage{url}

\usepackage[capitalize]{cleveref}
\usepackage{orcidlink}
\usepackage{amsthm}
\usepackage{enumitem}

\newtheorem{remark}{Remark}

\graphicspath{{./}}

\renewcommand{\cite}{\citep}

\def\eg{\emph{e.g.}}
\def\ie{\emph{i.e.}}

%% file: defs.tex
\newcommand{\token}{{x}}
\newcommand{\E}{\mathbb{E}} 
\newcommand{\R}{\mathbb{R}} 
\newcommand{\Man}{\mathcal{M}}

\newcommand{\T}{\mathcal{T}}
\newcommand{\TM}{\mathcal{T}\mathcal{M}}

\newcommand{\Loss}{\mathcal{L}} %
\newcommand{\netpars}{w}

\newcommand{\Unif}{\mathcal{U}}  %

\newcommand{\Exp}{\mathrm{{Exp}}}
\newcommand{\Log}{\mathrm{{Log}}}

\newcommand{\PM}{\mathcal{P}(\Man)}
\newcommand*\diff{\mathop{}\!\mathrm{d}}

\DeclareMathOperator{\grad}{\operatorname{grad}}
\newcommand\norm[1]{\lVert#1\rVert}

\DeclareMathOperator*{\argmin}{arg\min}

\crefname{equation}{eq.}{eq.}
\Crefname{equation}{Eq.}{Eq.}
\crefname{theorem}{thm.}{thms.}
\Crefname{Theorem}{Thm.}{Thms.}
\crefname{conjecture}{conj.}{conjs.}
\Crefname{Conjecture}{Conj.}{Conjs.}
\crefname{proposition}{prop.}{props.}
\Crefname{proposition}{Prop.}{Props.}
\crefname{definition}{dfn.}{dfn.}
\Crefname{definition}{Dfn.}{Dfn.}
\crefname{remark}{remark}{remark}
\Crefname{Remark}{Remark}{Remark}
\crefname{algorithm}{Alg.}{Alg.}
\Crefname{algorithm}{Alg.}{Alg.}

\newtheorem{dfn}{Definition}

\renewcommand{\paragraph}[1]{{\vspace{0.3mm}\noindent \bf #1}.}

\newacro{NVS}[NVS]{Novel View Synthesis}
\newacro{RFM}[RFM]{Riemannian Flow Matching}
\newacro{CD}[CD]{Chamfer Distance}
\newacro{VGGT}[VGGT]{Video Geometry Generative Transformer}
\newacro{VAE}[VAE]{Variational Autoencoder}
\newacro{RCFM}[RCFM] {Riemannian Conditional}
\newacro{SA}[SA]{Self-Attention}
\newacro{CA}[CA]{Cross-Attention}
\newacro{RoPE}[RoPE]{Rotary Position Embeddings}
\newacro{ProPE}[ProPE]{Projective Positional Encoding}
\newcommand{\damb}{d_{\mathrm{amb}}}

%% file: sec/0_abstract.tex
\begin{abstract}
Geometric foundation models such as the Visual Geometry Grounded Transformer (VGGT) provide strong 3D priors from unposed images, but operate deterministically and cannot generate plausible geometry beyond what the observations directly support. We bridge geometric foundation models and generative 3D modeling by performing flow matching directly in their latent space, leveraging learned 3D priors without committing to explicit representations such as Gaussians, meshes, or video-VAE latents. This requires respecting the latent geometry: LayerNorm places the decoder inputs on a product of zero-mean hyperspheres, and treating this space as unconstrained Euclidean substantially degrades flow matching. We therefore formulate Riemannian flow matching on this product manifold, conditioned on unposed context views and a target camera pose, keeping generated tokens within the geometry expected by the frozen decoder. Our main experiments use VGGT, while additional experiments with Depth Anything 3 show that the same formulation transfers to a different independently pretrained 3D backbone without modification. On RealEstate10K, ScanNet++, and ETH3D, our method achieves strong results against recent scene-generation baselines in both novel-view appearance and aggregated 3D geometry, establishing latent-space flow matching on geometric foundation models as a viable paradigm for 3D generation. Project page: \url{https://lisaweijler.github.io/geometry-grounded-rfm/}.

\end{abstract}

%% file: sec/1_intro.tex
\section{Introduction}
\label{sec:intro}

Inferring the complete 3D scene geometry from partial RGB observations is crucial for robotics, AR/VR, and autonomous driving.
Agents must not only understand the observed 3D scene, but also reason about what lies outside their field of view before they can safely plan through it.
This task, therefore, combines two different problems.
Where the input views provide evidence, the scene is determined; where they do not, many plausible geometries and appearances remain consistent with the observations.
The unobserved part of the task is therefore ill-posed in the strict sense: it admits a set of possible solutions rather than a single one.

From modern Novel View Synthesis (NVS) methods, such as NeRF~\cite{mildenhall2021nerf} and 3D Gaussian Splatting~\cite{kerbl20233d}, to recent feed-forward geometric models, such as DUSt3R~\cite{wang2024dust3r} and VGGT~\cite{wang2025vggt}, 3D reconstruction has made remarkable progress, producing high-quality renderings and accurate scene reconstructions from sets of input images.
However, 3D reconstruction methods address only part of the problem: they faithfully recover only the part of the scene supported by the input observations.

To obtain a complete scene from partial observations, recent methods therefore incorporate a generative mechanism rather than relying solely on reconstruction.
Due to the scarcity of 3D data, many of these methods rely on multi-view~\cite{Nair_2025_difftransf} or view-conditioned~\cite{Liu_2023_ICCV, elata2025novel,jang2025dtnvsdiffusiontransformersnovel} generative models, or leverage video generative models~\cite{liu2024reconxreconstructscenesparse,wang2024motionctrl,yu2025viewcrafter}.
However, such approaches often lack explicit 3D priors: they treat 3D reconstruction as a post-processing step and require geometric knowledge to be learned implicitly from large datasets via costly training, often resulting in multi-view inconsistencies.
More recent works have explored explicitly incorporating 3D structure, for instance, through depth-conditioning~\cite{seo2024genwarp,Ren_2025_gen3c,chen2025scenecompleter}, adding depth maps as a training objective~\cite{lu2025matrix3d}, or coupling generative models with computationally expensive reconstruction pipelines~\cite{wu2024reconfusion}.
Yet, these approaches still require costly training or reconstruction strategies.

Feed-forward geometric foundation models, such as VGGT~\cite{wang2025vggt}, provide strong geometric priors that can be exploited within generative frameworks.
A group of concurrent works has recognized this opportunity, pairing video generative models with the latent space of geometric foundation models~\cite{wu2025geometryforcing,huang2026gen3r}, or designing generative approaches directly within that latent space~\cite{jang2026repurposing}.
Operating directly in the native representation space of geometric foundation models is attractive because it allows generative models to directly leverage the strong 3D priors without first mapping them into a separate generative latent space. However, existing approaches treat this space as Euclidean and do not account for its underlying geometric structure, which can degrade performance.

In this work, inspired by generative approaches operating on structured latent spaces~\cite{zheng2026diffusion,kumar2026learning}, we propose a generative geometry-grounded foundation model that operates directly in the latent space of feed-forward reconstruction models.
However, naively applying standard generative techniques, such as flow matching~\cite{lipman2023flow}, to this space without accounting for its geometry can reduce generation quality.
To address this, we identify the latent representation of VGGT~\cite{wang2025vggt} and related models as a product manifold of four hyperspheres.
Building on~\cite{chen2024flow}, we then introduce a \emph{conditional Riemannian Flow Matching} (RFM) framework tailored to this latent geometry.
Our main experiments use VGGT, while additional experiments with Depth Anything 3 (DA3) demonstrate that the same formulation transfers to a different independently pretrained 3D backbone without modifying the manifold construction or generative model.
Our model is conditioned on a set of partial, unposed observations and a target camera pose.
By leveraging frozen decoding heads, we reconstruct depth and point maps directly from the generated latent codes.
Additionally, we train an RGB decoder for VGGT, enabling full RGB reconstruction.
We evaluate our framework on sparse-view reconstruction, demonstrating that it generates plausible 3D scenes while benefiting from the strong geometric priors encoded in the underlying reconstruction model.

In summary, \textbf{our contributions are}:
\begin{itemize}[leftmargin=*]
  \item We address 3D reconstruction and NVS from a generative perspective with a flow matching model operating \emph{directly} in the latent space of frozen geometric foundation models, thereby leveraging the strong 3D priors acquired during large-scale pre-training without first mapping them into a separate generative latent space.
  \item We realize this approach through a context- and pose-conditioned RFM model tailored to the latent structure of geometric foundation models. We instantiate it primarily with VGGT~\cite{wang2025vggt} and further demonstrate transfer to DA3, showing that the formulation is not tied to one backbone.
  \item Our extensive evaluation on RE10K, ScanNet++, and ETH3D demonstrates improved geometric reconstruction and generation while preserving high-quality RGB inference.
\end{itemize}

Our formulation is fully general with respect to the conditioning configuration: it is order-invariant over context views, handles as few as a single context view, and generates one target view at a time. This avoids the trajectory and adjacency assumptions of video-diffusion models.

%% file: sec/2_related_work.tex
\vspace{-2mm}
\section{Related Work}
\label{sec:rw}
\vspace{-2mm}

\paragraph{Multi-view 3D reconstruction}
Traditional multi-view reconstruction pipelines decompose the problem into feature matching, structure-from-motion, and dense reconstruction~\cite{schoenberger2016sfm, schoenberger2016mvs}, while neural approaches such as NeRF~\cite{mildenhall2021nerf, barron2021mip, muller2022instant} and Gaussian Splatting~\cite{kerbl20233d,huang20242d} bypass these steps via differentiable rendering at the cost of expensive per-scene optimization. Feed-forward variants generalize across scenes through explicit 3D representations~\cite{yu2021pixelnerf,chen2021mvsnerf,charatan2024pixelsplat,chen2024mvsplat,xu2025depthsplat} or geometry-free transformer architectures~\cite{jin2025lvsm,kim2026svsm}, but still require camera poses.
Geometric foundation models~\cite{wang2024dust3r,leroy2024grounding,wang2025vggt, lin2025depth} eliminate this requirement by jointly inferring geometry and camera parameters from uncalibrated images; VGGT~\cite{wang2025vggt} unifies camera estimation, dense geometry, and point correspondence in a single feed-forward transformer.
Our work exploits VGGT's structured latent space for generative modeling while preserving its geometric priors.

\paragraph{Reconstruction from limited (sparse) views}
Sparse-view reconstruction addresses scenarios where limited viewpoint coverage leaves geometry ambiguous and depth under-constrained.
Regularization-based approaches~\cite{niemeyer2022regnerf,jain2021putting,deng2022depth,kim2022infonerf} adapt per-scene optimization to sparse settings through geometric regularization, semantic consistency, or ray entropy minimization.
Feed-forward methods circumvent optimization through learned priors: some leverage multi-view feature matching to predict Gaussian splats~\cite{charatan2024pixelsplat,chen2024mvsplat,xu2025depthsplat,zhang2024gs}, while others learn scene-level priors to directly regress radiance fields~\cite{yu2021pixelnerf,chen2021mvsnerf}.
These deterministic approaches excel at reconstruction but cannot model distributions over plausible geometries. Our generative framework instead models 3D scenes probabilistically under sparse observations.

\paragraph{Scene generation}
For probabilistic scene modeling, methods leverage 2D generative priors or operate in 3D representation spaces.
Pixel-space methods synthesize novel views without explicit 3D structure, relying on implicit geometric learning through multi-view generation~\cite{Liu_2023_ICCV,Nair_2025_difftransf}, view-conditioned synthesis~\cite{elata2025novel,jang2025dtnvsdiffusiontransformersnovel}, or camera-controlled video diffusion~\cite{liu2024reconxreconstructscenesparse,yu2025viewcrafter,wang2024motionctrl,Zhang_2025_worldconsitent}.
To inject geometric priors while remaining in pixel space, methods align or augment video generators with geometric foundation models~\cite{wu2025geometryforcing, wan2026geoworld, dai2026fantasyworld, yang2026dreamworld}, condition generation on depth~\cite{seo2024genwarp,Ren_2025_gen3c,chen2025scenecompleter}, or tightly couple reconstruction and generation pipelines~\cite{wu2024reconfusion,gao2024cat3d,kong2025generative, lu2025prosplat}.
Other approaches explicitly generate geometry: Matrix3D~\cite{lu2025matrix3d} denoises RGB, depth, and camera rays jointly in a single multi-modal diffusion transformer, and GenRec~\cite{ccelen2026genrec} denoises RGB and scene coordinates while gating supervision by an observation mask, so that observed regions are regressed and unobserved ones generated.
Other work applies generative models to refine existing 3D reconstructions, using diffusion~\cite{wei2025gsfix3d,yin2025gsfixer,wu2025genfusion} or flow matching~\cite{fischer2025flowr} on rendered views to remove artifacts and complete occluded regions, or directly generates 3D primitives such as Gaussians~\cite{zhou2024diffgs,wewer2024latentsplat}.
Closest to us, Gen3R~\cite{huang2026gen3r} also builds on geometric foundation model tokens, but treats VGGT tokens as Euclidean vectors and maps them into the latent space of a pretrained video generative model, where generation takes place.
In contrast, we generate in VGGT's own token space and employ Riemannian flow matching to respect its hyperspherical geometry.

\paragraph{Latent spaces of deep networks}
Stable Diffusion~\cite{rombach2022high} moves generation from pixels to latent space through compressed VAE spaces.
Representation Autoencoders~\cite{zheng2026diffusion} instead pair frozen pretrained encoders such as DINOv2 with learned decoders, preserving semantically rich, high-dimensional representations.
Independently of our work, \citet{kumar2026learning} identify hyperspherical structure in DINOv2 embeddings and apply Riemannian flow matching to avoid the interference caused by Euclidean paths through the sphere's interior.

We observe that VGGT's latent space similarly forms a product manifold of hyperspheres and employ Riemannian flow matching for geometrically consistent 3D generation.
Recent work further supports operating in geometric-foundation-model latents: VGGT-World~\cite{sun2026vggtworld} forecasts frozen VGGT tokens for world modeling and reports that velocity-prediction flow matching collapses in this space, Geometry Latent Diffusion (GLD)~\cite{jang2026repurposing} repurposes DA3 and VGGT features as a multi-view diffusion latent space, and OneWorld~\cite{gao2026oneworld} generates in the representation space of a fine-tuned $\pi^3$~\cite{wang2026pi}, adopting clean-sample prediction following~\citet{li2025jit}.
All three, however, treat these features as Euclidean; we instead identify and exploit their hyperspherical product structure.

\paragraph{Riemannian flow matching (RFM)}
Flow matching~\cite{lipman2023flow} learns continuous normalizing flows by regressing velocity fields.
RFM extends this to Riemannian manifolds, with applications including molecule generation~\cite{huguet2024sequence,bosese}, human pose estimation~\cite{nadar2026posed,he2024nrdf}, material generation~\cite{miller2024flowmm}, and, as discussed above, representation-encoder latents for image generation~\cite{kumar2026learning}.
We extend RFM to 3D scene generation, operating directly on VGGT's geometric latent manifold to enable probabilistic modeling of 3D geometry while preserving strong reconstruction priors.

%% file: sec/3_methods.tex
\vspace{-2mm}
\section{Method}

\label{sec:methods}

\begin{figure}[t]
    \centering
    \includegraphics[width=\linewidth]{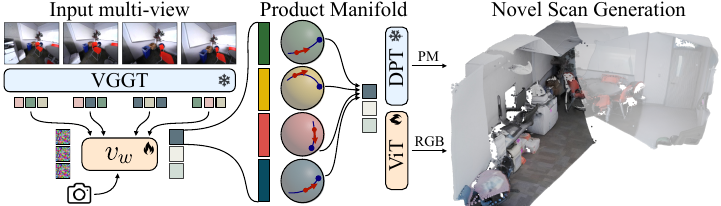}
    \caption{\textbf{Method overview.}
    Given a sparse set of unposed RGB observations, a frozen VGGT encoder extracts per-view latent tokens. A learned Riemannian flow matching model $v_w$ generates latent codes conditioned on the tokens of the input views and a target camera pose, operating directly on the product manifold of four zero-mean hyperspheres.  
    Frozen DPT decoding heads reconstruct point maps (PM) from the generated latents, while a trained RGB decoder produces full photometric reconstructions, enabling novel scan generation of complete, geometrically consistent 3D scenes. }
    \label{fig:overview}
\end{figure}
\vspace{-2mm}
Given a set of unposed RGB images $I = \{I^{(1)}, \dots, I^{(N)}\}$ of a scene and a target camera 
pose $\pi \in SE(3)$, our goal is to generate latent features that encode  the geometry and appearance of the scene from the target viewpoint, remaining multi-view consistent with the observed views. 
We cast this as a conditional generative modeling problem on the VGGT~\cite{wang2025vggt} encoder latent space: given context tokens $c = \{c^{(1)}, \dots, c^{(N)}\}$ extracted from context views $I$ and a target pose $\pi$, we model the conditional distribution $p(x_1 \mid c, \pi)$ on a Riemannian manifold $\Man$, where $x_1$ is the latent code corresponding to the target view. The generated $x_1$ is then decoded by VGGT's DPT heads into geometry (point clouds, depth) and, via an auxiliary head, into RGB pixels.
See \Cref{fig:overview} for an overview.

Next, we review RFM (\cref{sec:prelim-rfm}), characterize the VGGT latent space as a product of hyperspheres (\cref{sec:prelim-vggtlatent}), specialize conditional RFM to this setting (\cref{sec:vggt_latent_rfm}), and describe our architecture (\cref{sec:architecture}).

\vspace{-1mm}
\subsection{Riemannian Flow Matching}
\label{sec:prelim-rfm}
\vspace{-1mm}
We train an RFM model~\cite{chen2024flow} on a corpus of latent tokens
$\{\token^{(i)}\}$ living on a Riemannian manifold $\Man$. We briefly review the ingredients we use.

\paragraph{Flow and probability path}
A time-dependent flow on $\Man$ is a family of diffeomorphisms
$\{\psi_t \colon \Man \to \Man\}_{t \in [0,1]}$ obtained by integrating a
time-dependent vector field $u_t \in \Gamma(\TM)$:
\begin{equation}
    \label{eq:RODE}
    \tfrac{d}{dt}\,\psi_t(x) = u_t(\psi_t(x)), \qquad \psi_0(x) = x.
\end{equation}
The flow induces a probability path $p_t = [\psi_t]_{\#}\, p_0$ between a
reference distribution $p_0$ and the data distribution $p_1$, satisfying the
continuity equation $\partial_t p_t + \mathrm{div}_g(p_t u_t) = 0$.

\paragraph{Riemannian Conditional Flow Matching}
Since $u_t$ is intractable, RFM regresses a neural velocity field
$v_\netpars(x_t, t) \in \T_{x_t}\Man$ against a tractable \emph{conditional} field
$u_t(x \mid x_1)$ that generates a per-sample path $p_t(x \mid x_1)$ with
$p_0(x \mid x_1) = p_0(x)$ and $p_1(x \mid x_1) \approx \delta_{x_1}(x)$. For
the geodesic distance $d_g$ and linear schedule $\kappa(t) = 1-t$, the
minimal-norm conditional field admits the closed form
\begin{equation}
    \label{eq:ut-grad-form}
    u_t(x \mid x_1) = \tfrac{1}{1-t}\,\Log_x(x_1),
\end{equation}
yielding the RCFM training objective~\cite{chen2024flow}:
\begin{equation}
    \label{eq:RFMloss}
    \mathcal{L}_\mathrm{RCFM} = \E_{t,\, p(x_0),\, p(x_1)}
    \bigl\lVert v_\netpars(x_t, t) - u_t(x_t \mid x_1) \bigr\rVert^2_{g_{x_t}},
\end{equation}
with $t \sim \Unif(0,1)$ and $x_t = \Exp_{x_0}(t \Log_{x_0}(x_1))$ the
geodesic interpolant. 
We give more background and detailed definitions for RCFM in~\cref{supp_sub:rfm}.

\subsection{The VGGT latent space as a product of hyperspheres}
\label{sec:prelim-vggtlatent}

Operating on the latent space of VGGT~\cite{wang2025vggt} provides a strong
3D prior that our model can rely on to generate consistent 3D geometry. We
define this latent space as the input features of the final DPT heads in
the VGGT architecture, \ie, the output of the VGGT encoder, avoiding
costly self- and cross-attention operations during decoding. Features in
this space take the form $\token = \bigoplus_{i \in \{4, 11, 17, 23\}} h_i$,
where $\oplus$ is concatenation and $h_i$ is the output of VGGT's $i$th
encoder block. Each token has dimension $4C$ with $C = 2048$,
yielding $\token \in \R^{8192}$. 
Operating directly on encoder features avoids the reconstruction loss of an 
additional auto-encoder. Yet encoder features are typically high-dimensional latent representations that are difficult to model with Euclidean velocity-based diffusion or flow matching~\cite{yao2025vavae,zheng2026diffusion,li2025jit}.
This difficulty is typically attributed to the manifold hypothesis: data lies on a low-dimensional manifold within ambient space, and Euclidean probability paths traverse large off-manifold regions.

\noindent\textbf{Explicit manifold structure.}
In our case, the manifold structure is not hypothesized but explicit.
Before $\token$ is processed by the RGB and DPT heads, each $h_i$ passes through a
separate LayerNorm. This first removes the per-token mean (orthogonal
projection onto the zero-mean hyperplane $H = \{z \in \R^C : \langle \mathbf{1}, z \rangle = 0\}$),
then rescales the result to a fixed norm $\sqrt{C}$, projecting it onto the
hypersphere
\begin{equation}
    \mathcal{S}^{C-2} = \{x \in \R^C : \|x\|_2 = \sqrt{C},\ \langle \mathbf{1}, x \rangle = 0\}
\end{equation}
\begin{wrapfigure}{r}{0.4\textwidth}
  \vspace{-0.35cm}
  \begin{center}
    \includegraphics[width=0.39\textwidth]{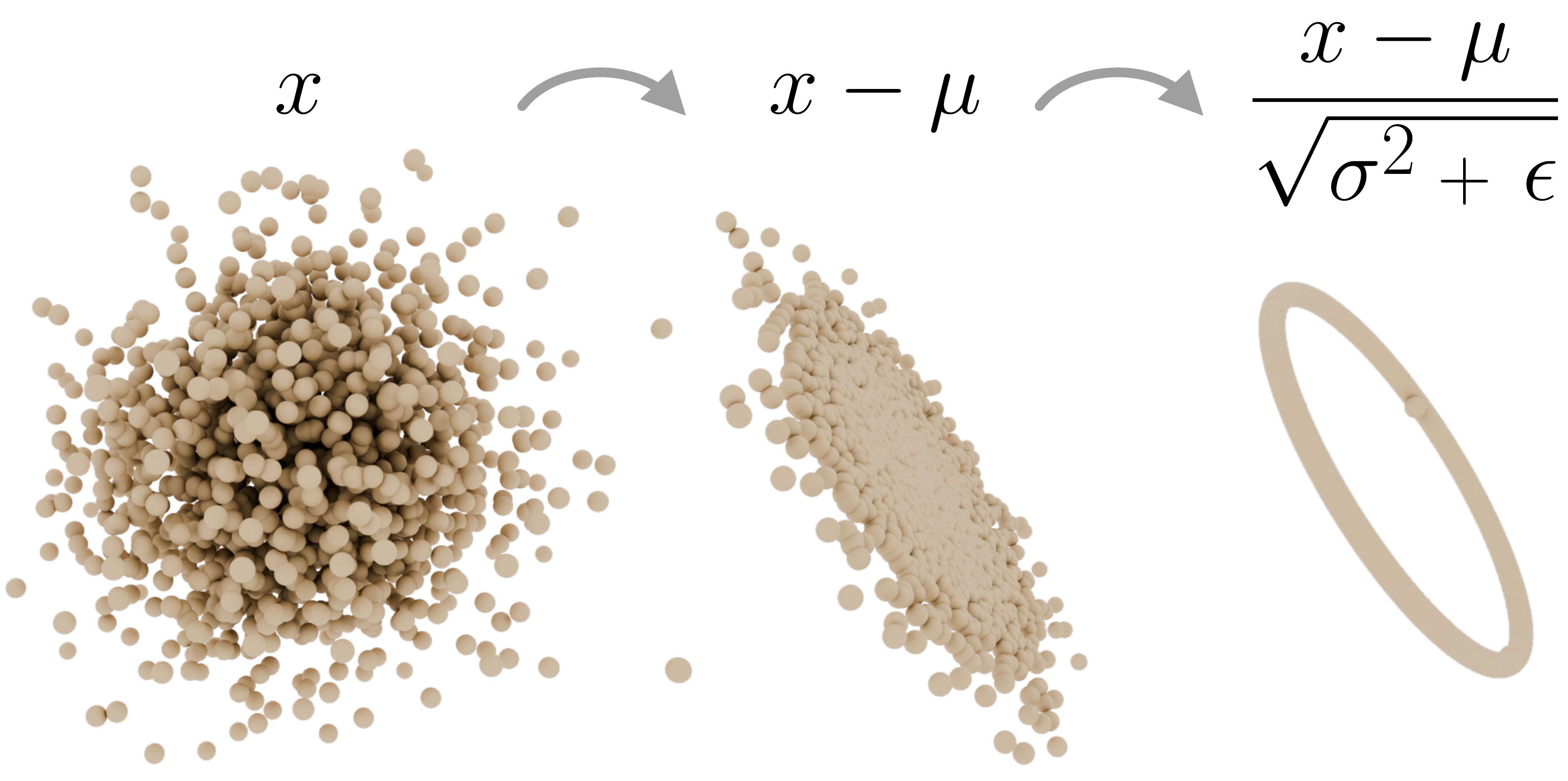}
  \end{center}
  \vspace{-0.5cm}
\end{wrapfigure}
embedded in $H$. The accompanying figure on the right illustrates this construction for
three dimensions. Consequently, operating directly on $\mathcal{S}^{C-2}$ incurs no loss of information relative to the post-LayerNorm representation, allowing us to fully exploit the learned affine parameters of LayerNorm and any downstream weights that act on these normalized features.

The full latent space is therefore the product manifold $\Man = \bigl(\mathcal{S}^{C-2}\bigr)^K,  K = 4$
of four zero-mean hyperspheres of radius $\sqrt{C}$.
Euclidean flow matching~\cite{lipman2023flow} ignores this structure: the prior, intermediate states, and velocity field are unconstrained in $\R^{4C}$, so trajectories can leave $\Man$ and reach the decoder off-manifold. Similar effects have been observed in other feature spaces~\cite{kumar2026learning}. We instead perform Riemannian flow matching directly on $\Man$, keeping the trajectory within the latent geometry expected by VGGT's frozen decoding heads by construction. We quantify the resulting gains in \cref{sec:exp_ablations}. For training, we rescale each
block to unit radius and undo the scaling before decoding.

\paragraph{Geometry of one block}
The tangent space at $x \in \mathcal{S}^{C-2}$ is
$\T_x \mathcal{S}^{C-2} = \{v \in \R^C : \langle v, x \rangle = 0,\ \langle \mathbf{1}, v \rangle = 0\}$,
i.e., vectors orthogonal to $x$ that also satisfy the zero-mean constraint.
With the induced metric, the exponential and logarithm maps (on the unit-radius sphere used for training) are
\begin{equation}
    \Exp_x(v) = \cos(\|v\|)\,x + \tfrac{\sin(\|v\|)}{\|v\|}\,v,
    \qquad
    \Log_x(y) = \tfrac{\theta}{\sin\theta}\bigl(y - \langle x, y\rangle x\bigr),
\end{equation}
with $\theta = \arccos(\langle x, y \rangle)$, and the tangent projection is
$\Pi_x(p) = p - \langle \mathbf{1}, p \rangle \,\mathbf{1}/C - \langle x, p\rangle x$
(enforcing the zero-mean and orthogonality constraints, respectively).

\paragraph{Product structure}
$\Man$ is endowed with the product metric
$d_\Man(x, x')^2 = \sum_{b=1}^{K} d_g\bigl(x^{(b)}, x'^{(b)}\bigr)^2$,
where $x^{(b)}$ denotes the $b$th block of $x$, \ie the normalized feature $h_i$ of the $b$th selected encoder block, and all maps act blockwise:
$\Exp_x = (\Exp_{x^{(1)}}, \dots, \Exp_{x^{(K)}})$, similarly for $\Log_x$ and
$\Pi_x$. \Cref{eq:RFMloss,eq:ConditionalRFMloss} therefore apply
independently to each of the four blocks. Hence, we describe the remaining
construction on a single sphere without loss of generality.

\subsection{Conditional RFM on VGGT latents}
\label{sec:vggt_latent_rfm}

For guided novel view generation, we condition on context tokens $c$ from observed views and a target camera pose $\pi \in SE(3)$.
In this conditional regime, our objective shifts from modeling the marginal target distribution $p(x_1)$ to modeling the conditional target distribution $p(x_1 \mid c, \pi)$. Consequently, the true time-dependent vector field driving the flow, $u_t(x \mid c, \pi)$, must be explicitly informed by both $c$ and $\pi$ to properly guide the probability paths across the product manifold $\Man$ toward valid, view-consistent 3D latents.

The conditional RCFM loss expands the expectation over the joint distribution
$p(x_1, c, \pi)$:
\begin{equation}
    \label{eq:ConditionalRFMloss}
    \mathcal{L}^{\mathrm{cond}}_\mathrm{RCFM}
    = \E_{t,\, p(x_0),\, p(x_1, c, \pi)}
    \bigl\lVert v_\netpars(x_t, t, c, \pi) - u_t(x_t \mid x_1) \bigr\rVert^2_{g_{x_t}}.
\end{equation}
Note that the target field $u_t(x \mid x_1)$ depends only on the endpoints
$x_0$ and $x_1$ alone: conditioning on $(c, \pi)$
shapes the joint distribution of $(x_1, c, \pi)$ from which training pairs
are sampled, but not the geodesic from $x_0$ to $x_1$ nor the
velocity along it.

At inference time, we sample an initial noise state $x_0 \sim \mathrm{Unif}(\Man)$ by sampling Gaussian noise and projecting it to  $\Man$. We integrate the learned vector field over the time domain $t \in [0, 1]$, where $t=0$ corresponds to the noise prior and $t=1$ to the data distribution. To ensure the integration path remains strictly on the manifold, we employ a Riemannian ODE solver (\eg, the Riemannian Euler method), iteratively updating the latent state via the exponential map:
\begin{equation}
    x_{t + \Delta t} = \Exp_{x_t}\left(\Delta t \cdot v_\netpars(x_t, t, c, \pi)\right).
\end{equation}

\subsection{Network architecture}
\label{sec:architecture}

We parameterize $v_\netpars(x_t,t,c,\pi)$ as a transformer that maps
manifold-valued inputs to tangent vectors. The four spherical components of
$x_t$ are concatenated and jointly projected to a working representation
$z\in\R^D$ with $D=768$, allowing information across components to mix from
the input. Context tokens $c$ are embedded separately into the same
working dimension. The backbone follows the pattern
$\text{SA-}(\text{CA-SA})^{8}$, comprising 9 self-attention (SA) blocks,
alternated with 8 cross-attention (CA) blocks.
All attention blocks use $12$ heads of dimension $64$, resulting in a model
with approximately $120$M parameters.

SA blocks use \ac{RoPE}~\cite{su2024roformer} to encode spatial patch structure and
inject the flow time $t$ through AdaLN modulation. CA blocks attend from the
latent representation $z$ to context tokens $c$, which serve as keys and
values, while the target pose $\pi$ is injected directly into cross-attention
through \ac{ProPE}~\cite{li2025cameras}.

A final AdaLN-modulated LayerNorm is followed by a dense linear map back to
the ambient space $\R^{4C}$. Its output is partitioned into the four latent blocks
and projected onto the corresponding tangent spaces via $\Pi_{x_t}$, yielding
the valid velocity
$v_\netpars(x_t,t,c,\pi)\in\T_{x_t}\Man$.
For RGB prediction, we additionally train a ViT head on ground-truth VGGT
latent tokens to decode the generated representation into RGB colors, enabling
novel view generation. Further implementation and training details of our flow model are given in~\cref{app:training_details} and details on the RGB head in~\cref{supp:rgb_head}.
\Cref{fig:overview} provides an overview of our method.

%% file: sec/4_experiments.tex
\section{Experiments}

\label{sec:experiments}
The generative nature of our approach enables plausible geometry and appearance to be generated for unseen regions, while faithfully preserving those of observed regions. 
We evaluate novel-view RGB and depth generation (\cref{sec:results_nvs}), aggregated
3D scene reconstruction (\cref{sec:3dgen}), and the effect of Euclidean versus Riemannian latent geometry (\cref{sec:exp_ablations}). 
Additional experiments in the supplement evaluate transfer to Depth Anything 3 (\cref{app:da3exp}). We compare against deterministic reconstruction, diffusion-based generation, and recent generative methods built on geometric foundation-model features.

\paragraph{Datasets} The model is trained on the training splits of \emph{RealEstate10K (RE10K)}~\cite{zhou2018stereo} and \emph{ScanNet++}~\cite{yeshwanth2023scannetpp} (59{,}103 scenes, 150 equidistant frames each). The two are complementary: ScanNet++ captures dense indoor scenes with rich camera motion, while RE10K adds outdoor and large-scale indoor scenes with smoother motion and predominantly forward-facing viewpoints. We evaluate on the RE10K test and ScanNet++ validation splits and, for 3D scan generation, on the \emph{ETH3D}~\cite{eth3d} training split as out-of-distribution data (${\sim} 7\times$ physically larger than ScanNet++); the evaluation scenes are listed in \cref{app:eval_scenes}.

\paragraph{Baselines}
We compare against four recent methods covering complementary approaches to novel view generation. DepthSplat~\cite{xu2025depthsplat} is a deterministic reconstruction method based on an explicit 3D representation.
Matrix3D~\cite{lu2025matrix3d} is a joint multimodal diffusion model that generates RGB in an image-VAE latent space while denoising depth and ray maps directly in 2D. Gen3R~\cite{huang2026gen3r} builds on VGGT features but maps
them into the latent space of a pretrained video model for generation.
GLD~\cite{jang2026repurposing} operates directly in VGGT feature space using Euclidean
flow matching, making it the closest existing baseline to our latent-space formulation.
GLD nevertheless differs from our method in several architectural choices: it generates features from two VGGT encoder levels in a cascade, propagates the remaining features, and uses a substantially larger generative model. We
therefore isolate the effect of latent geometry separately in~\cref{sec:exp_ablations}, where Euclidean and Riemannian flow matching use exactly the same architecture and training procedure.
Finally, all compared baselines use ground-truth context camera poses, whereas
our method operates on unposed context views. Additional baseline details are
provided in \cref{app:baselines}.

\paragraph{Evaluation protocols} We evaluate one-sided and two-sided context setups following Gen3R~\cite{huang2026gen3r} with $N \in \{1,2,3,4\}$ context views: one-sided uses the first $N$ images of a sequence as context and the last as target, two-sided uses the middle image as target with $N/2$ context views on each side (\cref{fig:sequence_layout}). DepthSplat requires at least two context views and Gen3R one or two, depending on the setting. We report two overlap regimes between target and context, \emph{high} (${\sim}60\%$ for ScanNet++, 95\% for RE10K) and \emph{low} (${\sim}35\%$ and 88\%). For 3D scan generation we use the two-sided low-overlap setting on ScanNet++ and context and target frames of ETH3D with ${\sim}40\%$ overlap. The point maps of the context and the two enclosed target views are fused into one scene cloud, aligned to ground truth with the Umeyama algorithm~\cite{umeyama1991}, and both are downsampled to $20$\,k points by farthest point sampling. All results use the Riemannian Euler ODE solver with 20 steps (${\sim}0.7$\,s per sample on an RTX~A6000; step-count ablation in \cref{supp_sub:odesteps}). Resolution unification, overlap categories, depth scale and point alignment are detailed in~\cref{app:eval_protocol}.

\paragraph{Metrics}
Following prior work~\cite{huang2026gen3r,xu2025depthsplat,wewer2024latentsplat,wang2025vggt}, we report PSNR, SSIM~\cite{ssim}, LPIPS~\cite{lpips}, and FID~\cite{fid} for RGB, RMSE, absolute relative error (ARel), $\delta{<}1.25$, and FID for depth, and Accuracy, Completeness, and Chamfer distance for point maps. On ScanNet++, where ground-truth (GT) depth is available, a per-target visibility mask separates the regions observed in the context views from unobserved ones: FID is computed on the prediction inpainted into the unobserved region of the GT frame, all other metrics on observed regions only, so that reconstruction and the generation of plausible unobserved content are evaluated separately (\cref{app:rec_gen}).

\input{sec/tab_geometry}
\input{sec/tab_nvs_rgb}

\subsection{Novel view generation}
\label{sec:results_nvs}

\begin{figure}[t]
    \centering
    \includegraphics[width=\linewidth]{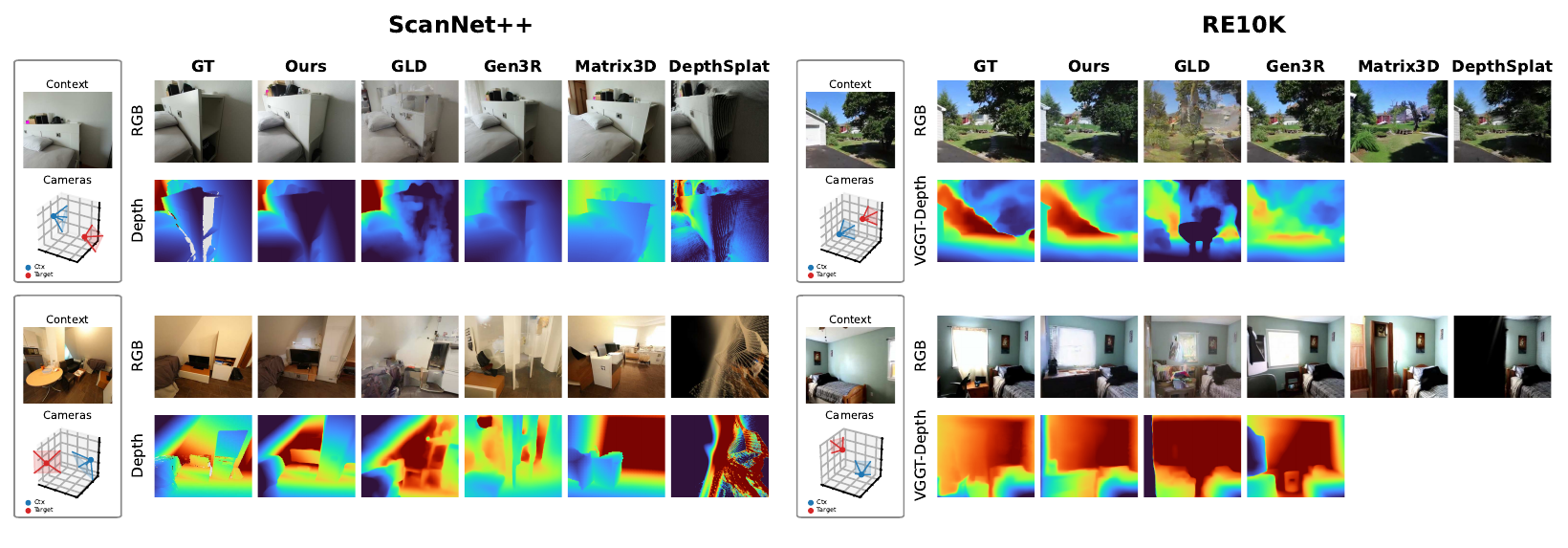}
    \caption{\textbf{Qualitative novel view generation on ScanNet++ (left) and RE10K (right).} For each scene, the columns show the target RGB and depth for the GT and all methods. On RE10K, depth is evaluated against VGGT pseudo-depth and is therefore shown only for the methods that share this backbone. Reconstruction-based DepthSplat leaves holes and streak artifacts where the target leaves its reconstructed support, GLD renders blurred or ghosted appearance, and Matrix3D and Gen3R produce plausible images with coarse or inconsistent depth, whereas our model recovers both appearance and geometry close to the ground truth.}
    \label{fig:nvs_qualitative}
\end{figure}
\paragraph{Depth prediction} \Cref{tab:nvs_depth} reports depth prediction on ScanNet++ and RE10K. ScanNet++ provides metric depth ground truth, while RE10K is evaluated against VGGT pseudo-depth and is therefore compared only with Gen3R and GLD, which likewise use VGGT as their geometric prior. On ScanNet++, our method achieves the best performance across all metrics, context counts, overlap regimes, and both one-sided and two-sided settings. The gains are largest in the more challenging one-sided, low-overlap regime, where larger camera offsets require greater extrapolation beyond the observed scene. In this setting, our method reduces RMSE by $28$-$45\%$ and ARel by $34$-$54\%$ relative to the strongest baseline for each context count, while improving $\delta$ by $0.11$-$0.28$ in absolute terms. 
On RE10K, where depth is evaluated against pseudo ground truth, we outperform both generative baselines in the one-sided setting, while GLD is better in the two-sided one.
Across both datasets, our largest gains occur in the more extrapolative settings, where the target is less constrained by the observations. Notably, our method achieves these results without requiring poses for the context views, unlike all competing baselines, and with a substantially smaller flow model than GLD (${\sim}120$M vs.\ ${\sim}1.3$B parameters).

\paragraph{RGB prediction}
\Cref{tab:nvs_rgb} reports RGB novel view generation
results on ScanNet++ and RE10K. On ScanNet++, our method achieves the
strongest results throughout the one-sided setting and under low overlap, with
the advantage again increasing for larger camera offsets similarly to depth. With a
single context view in the low-overlap regime, we improve PSNR from $13.86$ to $17.61$ and FID from $76.18$ to $63.86$ over the strongest competing values. The same trend holds with additional context and in the two-sided setting.
Only in the easiest two-sided, high-overlap configuration with $N{=}4$
DepthSplat obtains higher PSNR and SSIM (GLD also higher SSIM), while our method retains better LPIPS and FID. This is consistent with reconstruction-based methods
being particularly effective when the target is densely bracketed by observed
views, whereas our generative formulation is most beneficial when substantial content must be extrapolated beyond the observations.
On RE10K, where the smoother camera motion makes the task more interpolation-dominated, the differences are smaller. In the one-sided setting, our method achieves the strongest PSNR and LPIPS across all context counts and is best or competitive in SSIM and FID. In the two-sided
setting, DepthSplat and the other posed baselines benefit strongly from the
dense interpolation setting and achieve higher reconstruction metrics. 
Overall, the RGB results complement the geometric evaluation: the largest gains occur in the more extrapolative ScanNet++
regimes, while differences narrow when the target is well supported
by nearby context views.

\Cref{fig:nvs_qualitative} provides qualitative examples for depth and RGB across these settings, highlighting the improved geometric consistency of our predictions, particularly under large camera offsets and limited context.

\subsection{3D scan generation}
\label{sec:3dgen}
To measure not only 3D reconstruction and generation quality but also the consistency between generated and observed frames, we evaluate the point maps of the context frames jointly with those of their enclosed target views. \Cref{tab:3d_eval} reports the resulting fused-point-cloud metrics on ScanNet++ and ETH3D. On ScanNet++, our method achieves the best performance across all metrics, reducing the Chamfer distance from $13.24$\,cm to $11.43$\,cm relative to the strongest baseline, GLD, by improving both accuracy and completeness.

On the out-of-distribution ETH3D benchmark, our method remains competitive despite the substantially larger physical scene scale. GLD obtains the best accuracy and slightly lower Chamfer distance ($19.40$ vs.\ $20.36$\,cm), whereas our method achieves the best completeness ($14.76$ vs.\ $19.28$\,cm). The scale-normalized relative error remains similarly low for both methods ($0.69\%$ vs.\ $0.74\%$), indicating that our generated geometry transfers well to out-of-distribution scenes.

\begin{figure}[t]
    \centering
    \includegraphics[width=\linewidth]{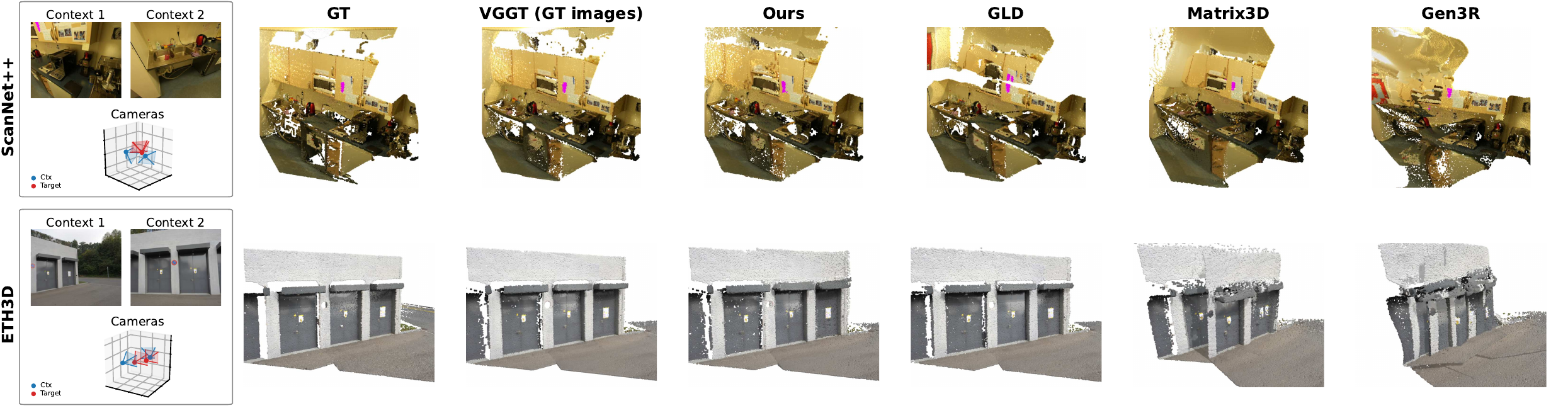}
    \caption{\textbf{Qualitative 3D scan generation on ScanNet++ (top) and ETH3D (bottom).} The columns show the point maps of the context and generated target views fused into a single scene, for the GT, for VGGT applied to the GT target images as an upper bound, and for all methods. Our model and GLD stay close to the reference geometry, whereas Matrix3D and Gen3R produce misaligned or duplicated structure.}
    \label{fig:qualitative_3d}
\end{figure}

\Cref{fig:qualitative_3d} qualitatively illustrates these results. Fusing the generated target point maps with those of the context frames yields coherent, well-aligned 3D scans, whereas competing methods exhibit larger geometric
distortions or incomplete structure.
Additional qualitative results for novel view generation and 3D scan generation are shown in \cref{fig_app:nvs_scannet_qualitative,fig_app:nvs_re10k_qualitative,fig_app:qualitative_3d}.

\subsection{Euclidean vs.\ Riemannian flow matching}
\label{sec:exp_ablations}

We directly ablate the effect of latent geometry by comparing Euclidean and Riemannian velocity flow matching with matched architecture, data, and training schedule. For the Euclidean baseline, latent channels are standardized using
training-set statistics and flow matching is performed in the resulting unconstrained Euclidean space. Before decoding, we invert the standardization transform. Our model instead defines the flow on $\Man$ and keeps sampling on it.

\begin{wrapfigure}{r}{0.35\textwidth}
  \vspace{-6pt}
  \centering
  \includegraphics[width=0.29\textwidth]{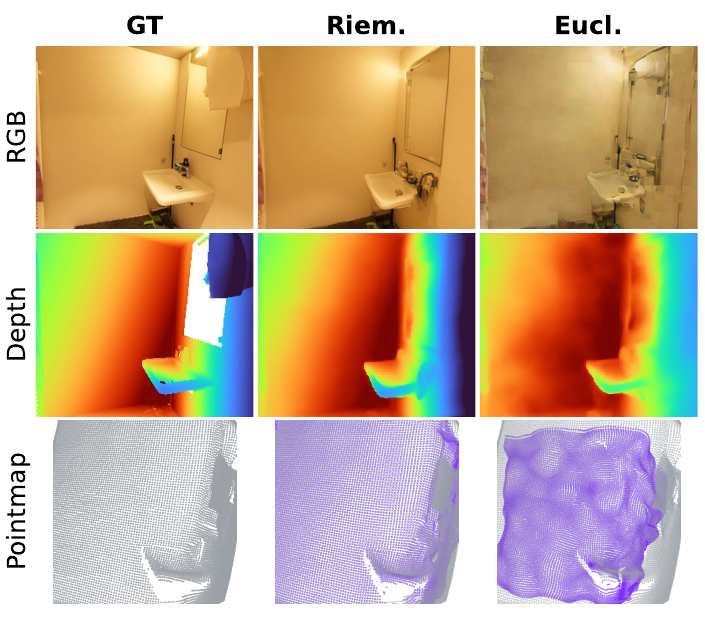}
  \vspace{-6pt}
  \caption{Eucl. vs.\ Riem. flow on a ScanNet++ scene.}
  \label{fig:eucl_vs_riem}
  \vspace{-10pt}
\end{wrapfigure}

As shown in \cref{tab:euclidvsriemannian} and \cref{fig:eucl_vs_riem} (more examples in \cref{fig_app:riem_vs_eucl_qual}), respecting the latent geometry substantially improves both RGB and depth generation. On the two-sided, low-overlap ScanNet++ setting with $N{=}4$ context views, RGB PSNR increases by
$1.41$\,dB and SSIM by $0.103$, while LPIPS and FID decrease by $41.8\%$ and $23.9\%$, respectively. For depth, RMSE and ARel decrease by $21.9\%$ and $43.8\%$, $\delta$ increases from $0.775$ to $0.907$, and FID decreases by
$8.9\%$. The same trend holds across all context and overlap configurations (\cref{app:eucl_vs_riemann}).

\textbf{Why the latent geometry matters.}
Euclidean flow is harder for two related reasons. Its straight-line transport contains radial motion, and the optimal velocity $u^\star(x,t)=\bigl(\mathbb{E}[x_1\mid x_t=x]-x\bigr)/(1-t)$ requires the model to reproduce the current state through the $-x/(1-t)$ term, which near $t=0$ is dominated by high-dimensional isotropic noise that cannot in general be preserved through the model's lower-dimensional representation. Riemannian flow restricts the velocity to the tangent space, removing both the radial component and this state-reconstruction burden.
The learned trajectories show the difference: Euclidean samples terminate at $\lVert x_1\rVert/R=1.33$, where $R$ is the empirical radius of real feature blocks, while Riemannian samples stay at unit radius. In addition to this radial deviation, the Euclidean endpoint also deviates in the directions it occupies. To quantify it, we measure the off-subspace energy $E_{\mathrm{off}}$ (\cref{eq:off_subspace}), the fraction of centered token energy that lies outside the per-block principal subspace capturing $99\%$ of the real-token variance. Intuitively, it is the share of generated-token variation in directions that real decoder inputs never occupy. Real tokens have $E_{\mathrm{off}}=1\%$, compared with $33\%$ for Euclidean and $2\%$ for Riemannian generated tokens, and Riemannian endpoints are better aligned with the ground-truth tokens (cosine similarity $0.85$ versus $0.66$). Euclidean flow thus leaves the decoder-relevant manifold both in radius and in direction. Trajectory, capacity, and prediction-target analyses are provided in \cref{app:trajectory_analysis}.

\begin{table}[t]
  \centering
  \caption{\textbf{Ablation on flow geometry.} Euclidean vs.\ Riemannian flow matching, matched architecture and training (two-sided ScanNet++, $N{=}4$, low overlap). Best in \textbf{bold}.}
  \label{tab:euclidvsriemannian}
  \setlength{\tabcolsep}{6pt}
  \scriptsize
  \begin{tabular}{l cccc cccc}
    \toprule
    & \multicolumn{4}{c}{RGB} & \multicolumn{4}{c}{Depth} \\
    \cmidrule(lr){2-5} \cmidrule(lr){6-9}
    Method & PSNR\,$\uparrow$ & SSIM\,$\uparrow$ & LPIPS\,$\downarrow$ & FID\,$\downarrow$
           & RMSE\,$\downarrow$ & ARel\,$\downarrow$ & $\delta$\,$\uparrow$ & FID\,$\downarrow$ \\
    \midrule
    Eucl. & 16.98 & 0.553 & 0.232 & 39.00 & 0.292 & 0.178 & 0.775 & 28.68 \\
    Riem. & {\bf 18.39} & {\bf 0.656} & {\bf 0.135} & {\bf 29.69} & {\bf 0.228} & {\bf 0.100} & {\bf 0.907} & {\bf 26.14} \\
    \bottomrule
  \end{tabular}
\end{table}

%% file: sec/tab_geometry.tex
\begin{table}[t]
  \providecommand{\metrichead}[1]{{\fontsize{6pt}{7pt}\selectfont #1}}
  \centering
  \caption{\textbf{Geometry evaluation.} (a) Novel-view depth on ScanNet++ (metric ground truth) and RE10K (VGGT pseudo-depth; only VGGT-based baselines). (b) 3D scan generation: accuracy, completeness and Chamfer distance of fused context and target point maps, in cm. Best in \textbf{bold}.}
  \label{tab:geometry}
  \setlength{\tabcolsep}{1.4pt}
  \setlength{\aboverulesep}{0pt}\setlength{\belowrulesep}{0pt}
  \scriptsize
  \begin{minipage}[t]{\textwidth}
    \subcaption{Novel-view depth generation}
    \label{tab:nvs_depth}
  \end{minipage}\par\vspace{-4pt}
  \newsavebox{\geomleftbox}
  \sbox{\geomleftbox}{\resizebox{0.515\textwidth}{!}{%
    \begin{tabular}[t]{p{9pt}p{6pt}l@{\hspace{5pt}}cccc cccc}
      \toprule
      & & & \multicolumn{8}{c}{ScanNet++} \\
      \cmidrule(lr){4-11}
      & & & \multicolumn{4}{c}{\rule{0pt}{2.3ex}High overlap} & \multicolumn{4}{c}{Low overlap} \\
      \cmidrule(lr){4-7} \cmidrule(lr){8-11}
      & N & \rule{0pt}{2.3ex}Method & \metrichead{RMSE$\downarrow$} & \metrichead{ARel$\downarrow$} & \metrichead{$\delta$$\uparrow$} & \metrichead{FID$\downarrow$} & \metrichead{RMSE$\downarrow$} & \metrichead{ARel$\downarrow$} & \metrichead{$\delta$$\uparrow$} & \metrichead{FID$\downarrow$} \\
      \midrule
      \multirow{16}{*}{\rotatebox[origin=c]{90}{one-sided}} & \multirow{4}{*}{$1$}
      & \rule{0pt}{2.3ex}Gen3R     & 0.386 & 0.223 & 0.723 & 55.86 & 0.543 & 0.301 & 0.571 & 64.75 \\
      & & Matrix3D  & 0.473 & 0.277 & 0.707 & 54.30 & 0.608 & 0.319 & 0.537 & 69.71 \\
      & & GLD       & 0.431 & 0.253 & 0.671 & 60.84 & 0.604 & 0.310 & 0.534 & 72.31 \\
      & & Ours      & \textbf{0.223} & \textbf{0.118} & \textbf{0.891} & \textbf{45.99} & \textbf{0.299} & \textbf{0.138} & \textbf{0.852} & \textbf{53.55} \\[1.2pt]
      \cmidrule(lr){2-11}
      & \multirow{4}{*}{$2$}
      & \rule{0pt}{2.3ex}DepthSplat& 0.488 & 0.295 & 0.582 & 48.47 & 0.540 & 0.289 & 0.590 & 52.85 \\
      & & Matrix3D  & 0.390 & 0.221 & 0.751 & 46.87 & 0.516 & 0.255 & 0.658 & 59.89 \\
      & & GLD       & 0.280 & 0.144 & 0.844 & 48.44 & 0.442 & 0.210 & 0.707 & 62.39 \\
      & & Ours      & \textbf{0.207} & \textbf{0.102} & \textbf{0.914} & \textbf{39.66} & \textbf{0.283} & \textbf{0.121} & \textbf{0.882} & \textbf{49.80} \\[1.2pt]
      \cmidrule(lr){2-11}
      & \multirow{4}{*}{$3$}
      & \rule{0pt}{2.3ex}DepthSplat& 0.461 & 0.274 & 0.606 & 40.98 & 0.541 & 0.294 & 0.581 & 50.45 \\
      & & Matrix3D  & 0.351 & 0.189 & 0.787 & 38.80 & 0.468 & 0.235 & 0.679 & 54.82 \\
      & & GLD       & 0.253 & 0.129 & 0.865 & 38.40 & 0.407 & 0.188 & 0.745 & 58.45 \\
      & & Ours      & \textbf{0.211} & \textbf{0.100} & \textbf{0.915} & \textbf{32.89} & \textbf{0.282} & \textbf{0.120} & \textbf{0.876} & \textbf{46.58} \\[1.2pt]
      \cmidrule(lr){2-11}
      & \multirow{4}{*}{$4$}
      & \rule{0pt}{2.3ex}DepthSplat& 0.441 & 0.249 & 0.634 & 32.99 & 0.536 & 0.289 & 0.581 & 48.83 \\
      & & Matrix3D  & 0.341 & 0.175 & 0.802 & 29.98 & 0.453 & 0.228 & 0.689 & 52.23 \\
      & & GLD       & 0.226 & 0.117 & 0.893 & 27.77 & 0.390 & 0.179 & 0.770 & 53.15 \\
      & & Ours      & \textbf{0.217} & \textbf{0.102} & \textbf{0.907} & \textbf{24.73} & \textbf{0.280} & \textbf{0.118} & \textbf{0.876} & \textbf{44.04} \\[1.2pt]
      \midrule
      \multirow{9}{*}{\rotatebox[origin=c]{90}{two-sided}} & \multirow{5}{*}{$2$}
      & \rule{0pt}{2.3ex}DepthSplat& 0.457 & 0.258 & 0.630 & 30.72 & 0.526 & 0.298 & 0.557 & 53.30 \\
      & & Gen3R     & 0.344 & 0.185 & 0.786 & 28.50 & 0.512 & 0.279 & 0.585 & 51.82 \\
      & & Matrix3D  & 0.354 & 0.183 & 0.810 & 29.30 & 0.442 & 0.227 & 0.702 & 49.91 \\
      & & GLD       & 0.257 & 0.120 & 0.873 & 27.94 & 0.340 & 0.168 & 0.778 & 49.24 \\
      & & Ours      & \textbf{0.220} & \textbf{0.094} & \textbf{0.908} & \textbf{23.64} & \textbf{0.276} & \textbf{0.128} & \textbf{0.856} & \textbf{41.41} \\[1.2pt]
      \cmidrule(lr){2-11}
      & \multirow{4}{*}{$4$}
      & \rule{0pt}{2.3ex}DepthSplat& 0.378 & 0.201 & 0.713 & 20.37 & 0.453 & 0.263 & 0.607 & 35.73 \\
      & & Matrix3D  & 0.306 & 0.145 & 0.852 & 18.28 & 0.372 & 0.190 & 0.760 & 33.45 \\
      & & GLD       & 0.216 & 0.087 & 0.905 & 15.93 & 0.247 & 0.115 & 0.877 & 30.18 \\
      & & Ours      & \textbf{0.211} & \textbf{0.085} & \textbf{0.921} & \textbf{13.23} & \textbf{0.228} & \textbf{0.100} & \textbf{0.907} & \textbf{26.14} \\[1.2pt]
      \bottomrule
    \end{tabular}}}
  \begin{minipage}[t]{0.515\textwidth}
    \centering\usebox{\geomleftbox}
  \end{minipage}%
  \hfill
  \begin{minipage}[t][\dimexpr\ht\geomleftbox+\dp\geomleftbox\relax][s]{0.47\textwidth}
    \centering
    \resizebox{\linewidth}{!}{%
    \begin{tabular}[t]{p{9pt}p{6pt}l@{\hspace{5pt}}cccc cccc}
      \toprule
      & & & \multicolumn{8}{c}{RE10K} \\
      \cmidrule(lr){4-11}
      & & & \multicolumn{4}{c}{\rule{0pt}{2.3ex}High overlap} & \multicolumn{4}{c}{Low overlap} \\
      \cmidrule(lr){4-7} \cmidrule(lr){8-11}
      & N & \rule{0pt}{2.3ex}Method & \metrichead{RMSE$\downarrow$} & \metrichead{ARel$\downarrow$} & \metrichead{$\delta$$\uparrow$} & \metrichead{FID$\downarrow$} & \metrichead{RMSE$\downarrow$} & \metrichead{ARel$\downarrow$} & \metrichead{$\delta$$\uparrow$} & \metrichead{FID$\downarrow$} \\
      \midrule
      \multirow{9}{*}{\rotatebox[origin=c]{90}{one-sided}} & \multirow{3}{*}{$1$}
      & \rule{0pt}{2.3ex}Gen3R     & 0.248 & 0.217 & 0.700 & 46.94 & 0.271 & 0.240 & 0.664 & 45.58 \\
      & & GLD       & 0.275 & 0.257 & 0.650 & 42.22 & 0.303 & 0.289 & 0.607 & 43.34 \\
      & & Ours      & \textbf{0.176} & \textbf{0.161} & \textbf{0.842} & \textbf{32.46} & \textbf{0.203} & \textbf{0.176} & \textbf{0.813} & \textbf{34.18} \\[1.2pt]
      \cmidrule(lr){2-11}
      & \multirow{2}{*}{$2$}
      & \rule{0pt}{2.3ex}GLD       & 0.240 & 0.215 & 0.729 & 39.16 & 0.272 & 0.249 & 0.677 & 39.97 \\
      & & Ours      & \textbf{0.171} & \textbf{0.157} & \textbf{0.851} & \textbf{32.04} & \textbf{0.198} & \textbf{0.172} & \textbf{0.820} & \textbf{34.38} \\[1.2pt]
      \cmidrule(lr){2-11}
      & \multirow{2}{*}{$3$}
      & \rule{0pt}{2.3ex}GLD       & 0.215 & 0.184 & 0.773 & 37.13 & 0.246 & 0.211 & 0.736 & 38.73 \\
      & & Ours      & \textbf{0.168} & \textbf{0.153} & \textbf{0.859} & \textbf{31.81} & \textbf{0.195} & \textbf{0.170} & \textbf{0.827} & \textbf{34.32} \\[1.2pt]
      \cmidrule(lr){2-11}
      & \multirow{2}{*}{$4$}
      & \rule{0pt}{2.3ex}GLD       & 0.197 & 0.170 & 0.812 & 35.72 & 0.228 & 0.194 & 0.772 & 37.90 \\
      & & Ours      & \textbf{0.166} & \textbf{0.151} & \textbf{0.860} & \textbf{31.69} & \textbf{0.192} & \textbf{0.169} & \textbf{0.831} & \textbf{33.89} \\[1.2pt]
      \midrule
      \multirow{5}{*}{\rotatebox[origin=c]{90}{two-sided}} & \multirow{3}{*}{$2$}
      & \rule{0pt}{2.3ex}Gen3R     & 0.131 & 0.119 & 0.886 & 50.90 & 0.145 & 0.150 & 0.828 & 63.08 \\
      & & GLD       & \textbf{0.064} & \textbf{0.044} & \textbf{0.965} & \textbf{30.04} & \textbf{0.096} & \textbf{0.091} & \textbf{0.925} & \textbf{43.75} \\
      & & Ours      & 0.094 & 0.075 & 0.940 & 30.28 & 0.118 & 0.107 & 0.901 & 44.17 \\[1.2pt]
      \cmidrule(lr){2-11}
      & \multirow{2}{*}{$4$}
      & \rule{0pt}{2.3ex}GLD       & \textbf{0.043} & \textbf{0.026} & \textbf{0.986} & \textbf{24.19} & \textbf{0.055} & \textbf{0.040} & \textbf{0.974} & \textbf{33.02} \\
      & & Ours      & 0.082 & 0.061 & 0.951 & 28.57 & 0.094 & 0.083 & 0.931 & 39.13 \\[1.2pt]
      \bottomrule
    \end{tabular}}
    \vfill
    \subcaption{3D scan generation}
    \label{tab:3d_eval}
    \vspace{-2pt}
    \raisebox{\depth}{\resizebox{\linewidth}{!}{%
    \begin{tabular}[t]{l@{\hspace{5pt}}cccc cccc}
      \toprule
      & \multicolumn{4}{c}{\rule{0pt}{2.3ex}ScanNet++} & \multicolumn{4}{c}{ETH3D} \\
      \cmidrule(lr){2-5} \cmidrule(lr){6-9}
      \rule{0pt}{2.3ex}Method & \metrichead{Acc$\downarrow$} & \metrichead{Comp$\downarrow$} & \metrichead{CH$\downarrow$} & \metrichead{Rel\%$\downarrow$} & \metrichead{Acc$\downarrow$} & \metrichead{Comp$\downarrow$} & \metrichead{CH$\downarrow$} & \metrichead{Rel\%$\downarrow$} \\
      \midrule
      \rule{0pt}{2.3ex}Gen3R     & 25.65 & 26.24 & 25.95 & 3.66 & 48.64 & 40.27 & 44.45 & 1.65 \\
      Matrix3D  & 22.59 & 16.87 & 19.73 & 2.75 & 41.00 & 22.20 & 31.60 & 1.26 \\
      GLD       & 13.90 & 12.58 & 13.24 & 1.85 & \textbf{19.53} & 19.28 & \textbf{19.40} & \textbf{0.69} \\
      Ours      & \textbf{12.45} & \textbf{10.42} & \textbf{11.43} & \textbf{1.55} & 25.97 & \textbf{14.76} & 20.36 & 0.74 \\[1.2pt]
      \bottomrule
    \end{tabular}}}
  \end{minipage}
\end{table}

%% file: sec/tab_nvs_rgb.tex
\begin{table}[t]
  \providecommand{\metrichead}[1]{{\fontsize{6pt}{7pt}\selectfont #1}}
  \centering
  \caption{\textbf{Novel-view RGB generation on ScanNet++ (left) and RE10K (right).} Best in \textbf{bold}.}
  \label{tab:nvs_rgb}
  \setlength{\tabcolsep}{1.2pt}
  \setlength{\aboverulesep}{0pt}\setlength{\belowrulesep}{0pt}
  \scriptsize
  \begin{tabular}[t]{p{9pt}p{6pt}l@{\hspace{5pt}}cccc cccc}
    \toprule
    & & & \multicolumn{8}{c}{ScanNet++} \\
    \cmidrule(lr){4-11}
    & & & \multicolumn{4}{c}{\rule{0pt}{2.3ex}High overlap} & \multicolumn{4}{c}{Low overlap} \\
    \cmidrule(lr){4-7} \cmidrule(lr){8-11}
    & N & \rule{0pt}{2.3ex}Method & \metrichead{PSNR$\uparrow$} & \metrichead{SSIM$\uparrow$} & \metrichead{LPIPS$\downarrow$} & \metrichead{FID$\downarrow$} & \metrichead{PSNR$\uparrow$} & \metrichead{SSIM$\uparrow$} & \metrichead{LPIPS$\downarrow$} & \metrichead{FID$\downarrow$} \\
    \midrule
    \multirow{16}{*}{\rotatebox[origin=c]{90}{one-sided}} & \multirow{4}{*}{$1$}
    & \rule{0pt}{2.3ex}Gen3R     & 15.39 & 0.543 & 0.143 & 65.78 & 13.86 & 0.475 & 0.153 & 77.66 \\
    & & Matrix3D  & 14.78 & 0.535 & 0.159 & 63.09 & 13.21 & 0.477 & 0.159 & 76.18 \\
    & & GLD       & 13.97 & 0.509 & 0.189 & 82.72 & 13.25 & 0.463 & 0.170 & 98.40 \\
    & & Ours      & \textbf{18.99} & \textbf{0.677} & \textbf{0.085} & \textbf{49.73} & \textbf{17.61} & \textbf{0.620} & \textbf{0.086} & \textbf{63.86} \\[1.2pt]
    \cmidrule(lr){2-11}
    & \multirow{4}{*}{$2$}
    & \rule{0pt}{2.3ex}DepthSplat& 13.51 & 0.514 & 0.239 & 94.03 & 11.20 & 0.365 & 0.294 & 137.2 \\
    & & Matrix3D  & 15.95 & 0.572 & 0.151 & 51.57 & 14.33 & 0.515 & 0.154 & 70.52 \\
    & & GLD       & 16.10 & 0.601 & 0.149 & 60.22 & 14.50 & 0.527 & 0.156 & 84.30 \\
    & & Ours      & \textbf{19.13} & \textbf{0.681} & \textbf{0.096} & \textbf{42.16} & \textbf{17.76} & \textbf{0.630} & \textbf{0.095} & \textbf{58.86} \\[1.2pt]
    \cmidrule(lr){2-11}
    & \multirow{4}{*}{$3$}
    & \rule{0pt}{2.3ex}DepthSplat& 15.15 & 0.607 & 0.201 & 67.18 & 11.94 & 0.417 & 0.280 & 117.7 \\
    & & Matrix3D  & 16.82 & 0.599 & 0.145 & 41.55 & 14.56 & 0.528 & 0.158 & 67.16 \\
    & & GLD       & 16.93 & 0.639 & 0.147 & 45.90 & 15.04 & 0.553 & 0.154 & 74.46 \\
    & & Ours      & \textbf{19.35} & \textbf{0.685} & \textbf{0.105} & \textbf{35.30} & \textbf{17.78} & \textbf{0.632} & \textbf{0.102} & \textbf{55.10} \\[1.2pt]
    \cmidrule(lr){2-11}
    & \multirow{4}{*}{$4$}
    & \rule{0pt}{2.3ex}DepthSplat& 16.63 & 0.661 & 0.181 & 44.24 & 12.47 & 0.441 & 0.280 & 108.6 \\
    & & Matrix3D  & 17.93 & 0.629 & 0.142 & 30.43 & 14.56 & 0.532 & 0.167 & 62.34 \\
    & & GLD       & 17.86 & 0.681 & 0.142 & 31.99 & 15.37 & 0.573 & 0.154 & 67.42 \\
    & & Ours      & \textbf{19.59} & \textbf{0.691} & \textbf{0.113} & \textbf{25.67} & \textbf{17.73} & \textbf{0.632} & \textbf{0.110} & \textbf{51.86} \\[1.2pt]
    \midrule
    \multirow{9}{*}{\rotatebox[origin=c]{90}{two-sided}} & \multirow{5}{*}{$2$}
    & \rule{0pt}{2.3ex}DepthSplat& 17.50 & 0.647 & 0.166 & 41.02 & 12.56 & 0.469 & 0.270 & 92.42 \\
    & & Gen3R     & 17.27 & 0.596 & 0.159 & 34.77 & 14.00 & 0.498 & 0.214 & 67.73 \\
    & & Matrix3D  & 17.40 & 0.605 & 0.153 & 31.70 & 14.94 & 0.545 & 0.178 & 59.42 \\
    & & GLD       & 17.17 & 0.640 & 0.163 & 35.11 & 14.98 & 0.555 & 0.178 & 68.91 \\
    & & Ours      & \textbf{19.58} & \textbf{0.687} & \textbf{0.112} & \textbf{26.10} & \textbf{17.43} & \textbf{0.631} & \textbf{0.121} & \textbf{50.67} \\[1.2pt]
    \cmidrule(lr){2-11}
    & \multirow{4}{*}{$4$}
    & \rule{0pt}{2.3ex}DepthSplat& \textbf{20.31} & \textbf{0.730} & 0.143 & 20.45 & 16.34 & 0.599 & 0.224 & 44.62 \\
    & & Matrix3D  & 19.23 & 0.661 & 0.133 & 17.55 & 15.89 & 0.572 & 0.203 & 35.26 \\
    & & GLD       & 18.95 & 0.719 & 0.140 & 17.25 & 16.85 & 0.632 & 0.171 & 35.74 \\
    & & Ours      & 20.18 & 0.703 & \textbf{0.116} & \textbf{12.89} & \textbf{18.39} & \textbf{0.656} & \textbf{0.135} & \textbf{29.69} \\[1.2pt]
    \bottomrule
  \end{tabular}%
  \hspace{6pt}%
  \begin{tabular}[t]{cccc cccc}
    \toprule
    \multicolumn{8}{c}{RE10K} \\
    \cmidrule(lr){1-8}
    \multicolumn{4}{c}{\rule{0pt}{2.3ex}High overlap} & \multicolumn{4}{c}{Low overlap} \\
    \cmidrule(lr){1-4} \cmidrule(lr){5-8}
    \rule{0pt}{2.3ex}\metrichead{PSNR$\uparrow$} & \metrichead{SSIM$\uparrow$} & \metrichead{LPIPS$\downarrow$} & \metrichead{FID$\downarrow$} & \metrichead{PSNR$\uparrow$} & \metrichead{SSIM$\uparrow$} & \metrichead{LPIPS$\downarrow$} & \metrichead{FID$\downarrow$} \\
    \midrule
    \rule{0pt}{2.3ex}14.66 & 0.475 & 0.367 & \textbf{40.94} & 14.24 & 0.463 & 0.397 & \textbf{43.27} \\
    14.03 & 0.467 & 0.410 & 47.07 & 13.36 & 0.449 & 0.452 & 49.81 \\
    14.41 & 0.455 & 0.458 & 66.96 & 14.06 & 0.445 & 0.484 & 69.35 \\
    \textbf{15.80} & \textbf{0.511} & \textbf{0.327} & 47.20 & \textbf{15.19} & \textbf{0.496} & \textbf{0.356} & 49.98 \\[1.2pt]
    \cmidrule(lr){1-8}
    \rule{0pt}{2.3ex}10.39 & 0.321 & 0.537 & 148.1 & 9.63 & 0.277 & 0.591 & 169.0 \\
    14.17 & 0.470 & 0.401 & 47.65 & 13.64 & 0.458 & 0.436 & 49.69 \\
    15.09 & 0.488 & 0.408 & 57.29 & 14.52 & 0.471 & 0.444 & 60.30 \\
    \textbf{15.87} & \textbf{0.513} & \textbf{0.324} & \textbf{46.64} & \textbf{15.23} & \textbf{0.498} & \textbf{0.353} & \textbf{49.32} \\[1.2pt]
    \cmidrule(lr){1-8}
    \rule{0pt}{2.3ex}11.38 & 0.371 & 0.511 & 145.6 & 10.49 & 0.322 & 0.563 & 166.32 \\
    14.28 & 0.475 & 0.392 & 48.59 & 13.74 & 0.462 & 0.423 & 49.58 \\
    15.55 & 0.512 & 0.378 & 53.06 & 14.92 & 0.489 & 0.417 & 56.99 \\
    \textbf{15.92} & \textbf{0.514} & \textbf{0.321} & \textbf{45.92} & \textbf{15.29} & \textbf{0.500} & \textbf{0.350} & \textbf{49.02} \\[1.2pt]
    \cmidrule(lr){1-8}
    \rule{0pt}{2.3ex}11.62 & 0.385 & 0.507 & 146.6 & 10.79 & 0.339 & 0.556 & 162.6 \\
    14.53 & 0.484 & 0.378 & 47.94 & 13.93 & 0.469 & 0.413 & 50.64 \\
    15.86 & \textbf{0.531} & 0.360 & 49.60 & 15.24 & \textbf{0.508} & 0.395 & 53.44 \\
    \textbf{15.98} & 0.516 & \textbf{0.317} & \textbf{45.11} & \textbf{15.36} & 0.502 & \textbf{0.347} & \textbf{48.91} \\[1.2pt]
    \midrule
    \rule{0pt}{2.3ex}\textbf{24.61} & \textbf{0.833} & \textbf{0.092} & 28.99 & \textbf{20.54} & \textbf{0.724} & \textbf{0.192} & 61.53 \\
    21.27 & 0.718 & 0.113 & \textbf{25.84} & 18.02 & 0.598 & 0.214 & \textbf{47.81} \\
    19.69 & 0.659 & 0.155 & 43.36 & 17.28 & 0.578 & 0.236 & 63.06 \\
    19.68 & 0.684 & 0.174 & 39.58 & 17.93 & 0.604 & 0.251 & 60.56 \\
    18.44 & 0.592 & 0.195 & 40.83 & 16.83 & 0.539 & 0.265 & 57.03 \\[1.2pt]
    \cmidrule(lr){1-8}
    \rule{0pt}{2.3ex}\textbf{24.20} & \textbf{0.845} & \textbf{0.094} & 33.40 & \textbf{21.98} & \textbf{0.778} & \textbf{0.143} & 52.32 \\
    21.82 & 0.722 & 0.111 & 36.55 & 19.76 & 0.655 & 0.157 & 49.60 \\
    21.36 & 0.752 & 0.132 & \textbf{31.77} & 20.42 & 0.705 & 0.161 & \textbf{42.51} \\
    19.08 & 0.610 & 0.170 & 38.11 & 17.77 & 0.567 & 0.217 & 49.41 \\[1.2pt]
    \bottomrule
  \end{tabular}
\end{table}

%% file: sec/5_conclusion.tex
\section{Conclusion}
\label{sec:conclusion}

We introduce a generative framework for 3D scene modeling that performs flow
matching directly in the latent space of a frozen geometric foundation model,
VGGT, rather than committing to an explicit downstream representation such as
Gaussians, meshes, or video-VAE latents. The key observation is that VGGT's
post-LayerNorm tokens do not occupy a Euclidean space but a product manifold of four zero-mean hyperspheres, which standard flow matching ignores. Our
conditional Riemannian flow matching formulation respects this geometry,
keeping generated tokens on the valid data manifold required by the frozen
decoding heads. The resulting model is order-invariant in the context views,
operates from as few as a single unposed view, and requires only the relative
target pose, removing the trajectory and adjacency assumptions of
video-diffusion and explicit-reconstruction baselines. On ScanNet++ and RE10K
it improves over recent scene-generation methods in per-view geometry and
remains competitive in appearance, and on the 3D scan generation task it
produces coherent, well-aligned point clouds that generalize to the
out-of-distribution ETH3D setting. Together these results suggest latent-space
Riemannian flow matching on geometric foundation models is a viable paradigm
for generative 3D reconstruction.

\paragraph{Limitations and future work}
\label{sec:limitations}
Our current framework generates one target view at a time. The 3D scan experiment shows that jointly decoded frames
already remain mutually consistent, yet extending it to joint multi-view generation could improve global consistency across synthesized views.
More broadly, casting reconstruction as a flow on VGGT's latent manifold opens directions unavailable to a purely feed-forward model: sampling the flow yields a distribution over plausible scenes, offering a natural route to uncertainty estimation in unobserved regions, while the structure of the learned trajectories invites exploration for tasks such as multi-view alignment and scene editing.

\newpage

\subsubsection*{Acknowledgements and disclosures}
\paragraph{Funding transparency statement}
Funding in direct support of this work: Austrian Research Promotion Agency (FFG) project FO999922723; a TU Wien KUWI grant (kurzfristige wissenschaftliche Arbeiten und fachspezifische Kurse im Ausland) supporting LW's research visit at Imperial College London; UKRI Engineering and Physical Sciences Research Council (EPSRC) Future Leaders Fellowship MR/Y018818/1 (TB); compute from the UK AI Research Resource (AIRR), grant 0251-4584-0945-1, from the Austrian Scientific Computing (ASC) infrastructure, in particular the MUSICA system, project 201194, and from the EuroHPC Joint Undertaking, project EHPC-DEV-2025D11-012, on the EuroHPC supercomputer LEONARDO hosted by CINECA.

\paragraph{Competing interests}
The authors declare no competing financial interests related to this work.

%% file: sec/X_supp.tex
\newpage
\appendix
\section*{Appendix}

\renewcommand{\thetable}{\thesection.\arabic{table}}
\renewcommand{\thefigure}{\thesection.\arabic{figure}}
\numberwithin{table}{section}
\numberwithin{figure}{section}
\numberwithin{equation}{section}

\section{Theoretical background}
\label{supp:theory}

\subsection{Riemannian Flow Matching} 
\label{supp_sub:rfm}

\begin{dfn}[Riemannian Flow~\cite{chen2024flow}]
A time-dependent \textbf{flow} is a one-parameter family of diffeomorphisms $\{\psi_t:\Man\to\Man\}_{t{=}0}^1$
    defined by integrating instantaneous deformations represented by a time-dependent vector field $u_t \in \Gamma(\TM)$ on the tangent space (Riemannian flow-matching field). $\psi_t$ is defined by solving the following Riemannian ordinary differential equation (ODE) on $\Man$ over $t \in[0,1]$:
    \begin{equation}\label{eq:RODE_suppl}
        \frac{d}{d t} \psi_t(x)=u_t\left(\psi_t(x)\right), \quad \psi_0(x)=x.
    \end{equation}
\end{dfn}
\begin{dfn}[Probability path]
Let $\PM$ denote the space of probability distributions on $\Man$. A probability path $p_t:[0,1]\to\PM$ interpolates between two distributions $p_0,p_1\in\PM$ indexed by $t\in[0,1]$.
$p_t$ is said to be \textbf{generated} by $\psi_t$ if it \emph{pushes forward} $p_0:=p(x_0)$ to $p_1:=p(x_1)$ following $u_t$, \ie $p_t=[\psi_t]_{\#}(p_0)$. We define a smooth probability path between data $p_1$ and a reference $p_0$ as
\begin{align}
\label{eq:gpp}
p_t(x)=\int_{\Man} p_t(x \mid x_1)\,p_1(x_1) \,\diff V(x_1),
\end{align}
where $p_t(x \mid x_1)$ is the conditional probability path obtained by pushing the source forward along the conditional flow $\psi_t(\cdot \mid x_1)$ generated by the conditional vector field $u_t(x \mid x_1)$ defined below, \ie $p_t(\cdot \mid x_1)=[\psi_t(\cdot \mid x_1)]_{\#}\,p_0$, so that $p_0(\cdot \mid x_1)=p_0$ and $p_1(\cdot \mid x_1)$ concentrates at $x_1$. 
\end{dfn}
\begin{remark}[Velocity field]
    $p_t$ satisfies a continuity (Liouville) equation on the manifold: ${\partial_t p_t}+\mathrm{div}_g\left(p_t u_t\right)=0$, where $u_t$ is the \textbf{velocity field} transporting probability mass along the manifold, where $\mathrm{div}_g$ is the divergence on $\Man$.
\end{remark}
We are now ready to define Riemannian-FM (RFM).
\begin{dfn}[RFM]
Given a probability path $p_t$, subject to the boundary conditions $p_0=p_{\mathrm{source}}$ and $p_1=p_{\mathrm{target}}$, as well as an associated flow $\psi_t$, Riemannian flow matching learns a \emph{continuous normalizing flow} by directly regressing $u_t$ through a neural network $v_{\netpars}(x,t)$ parametrized by $w$.
\end{dfn}
\begin{dfn}[Riemannian Conditional FM]
    The vanilla RFM objective is intractable as we do not have access to the closed-form $u_t$ generating $p_t$. Instead, we regress $v_{\netpars}$ against a tractable \emph{conditional vector field} $u_t(x\mid x_1)$, generating a \emph{conditional probability path} $p_t(x\mid x_1)$ which can recover the target unconditional path by marginalization:
    \begin{equation}
        u_t(x) = \int_{\Man} u_t(x\mid x_1)\frac{p_t(x\mid x_1)p(x_1)}{p_t(x)}\diff \mathrm{V}_{x_1}.
    \end{equation}
\end{dfn}

\begin{dfn}[Generating conditional vector field]
RFM defines a vector field  $u_t\left(x \mid x_1\right)$ that generates $p_t(x \mid x_1)$ through a distance $d$ by enforcing $d\left(\psi_t\left(x \mid x_1\right), x_1\right)=\kappa(t) d\left(x, x_1\right)$. The minimal-norm conditional field is~\cite{chen2024flow}:
\begin{equation}
u_t\left(x \mid x_1\right)=\frac{d}{d t} \log \kappa(t) \frac{d\left(x, x_1\right)}{\left\|\nabla d\left(x, x_1\right)\right\|_g^2} \nabla d\left(x, x_1\right).
\end{equation}
For the geodesic distance $d:=d_g$ and $\kappa(t):=1-t$, $\left\|\nabla d_g\right\|_g=1$ and $d_g \nabla d_g=\nabla \frac{1}{2} d_g^2=-\Log _x\left(x_1\right)$, giving
\begin{equation}
u_t\left(x \mid x_1\right)=\frac{1}{1-t} \Log _x\left(x_1\right)
\end{equation}
\end{dfn}
With this choice of time scheduling 
Chen \& Lipman~\cite{chen2024flow} then define an explicit Riemannian conditional FM (RCFM) objective for learning as:   
\begin{equation}
\label{eq:RFMloss2}
\Loss_{\mathrm{RCFM}}=
\E_{t, p(x_1), p(x_0)} \left\lVert v_w(x_t, t) + d(x_0, x_1) \frac{\grad~d(x_t, x_1)}{\norm{\grad~d(x_t, x_1)}^2_g}\right\rVert ^2_g,
\end{equation}
whose gradient is the same as that of RFM. 
We use \cref{eq:RFMloss2} to train our model, \ie, to obtain the parameters $\netpars$ of $v_\netpars$. Here, $t~\in\Unif(0,1)$ and $d(\cdot,\cdot)$ is the geodesic distance.

\paragraph{Parallel-Transport Form of the RCFM Target}
In our implementation, we compute the conditional target velocity via
parallel transport rather than the gradient form of \cref{eq:RFMloss2}.
The two are equivalent on constant-speed geodesics, as we now show.

Recall that the conditional vector field at $x_t$ is
\begin{equation}
    u_t(x_t \mid x_1) = \frac{1}{1-t}\Log_{x_t}(x_1).
\end{equation}
Let $P_{x_0 \to x_t} \colon \T_{x_0}\Man \to \T_{x_t}\Man$ denote parallel
transport\footnote{Often used to compare tangent vectors at distinct points~\cite{boumal2023introduction}.} along the geodesic from $x_0$ to $x_t$ defined by
$x_t = \Exp_{x_0}(t \cdot \Log_{x_0}(x_1))$. Then
\begin{equation}
\label{eq:pt-equiv-supp}
    u_t(x_t \mid x_1)
    \;=\;
    P_{x_0 \to x_t}\bigl(\Log_{x_0}(x_1)\bigr).
\end{equation}

To see this, note that the conditional path is the geodesic
$\gamma(t)=\Exp_{x_0}(t\,\Log_{x_0}(x_1))$ and the target velocity is
$\dot\gamma(t)$. A geodesic is self-parallel, $\nabla_{\dot\gamma}\dot\gamma=0$,
so its velocity at time $t$ is the parallel transport of its initial velocity
$\Log_{x_0}(x_1)$; since $d_g(x_t,x_1)=(1-t)\,d_g(x_0,x_1)$, this is the same
vector as $\frac{1}{1-t}\Log_{x_t}(x_1)$.

We use \cref{eq:pt-equiv-supp} in our implementation because it is
well-conditioned for all $t$: the logarithm is evaluated once, between $x_0$
and $x_1$, which are far apart, and transported by a rotation in the plane
they span. The gradient form instead requires $\Log_{x_t}(x_1)$ between points
that coincide as $t\to1$, where both the $\arccos$ and the projection lose
precision, and then divides the result by $1-t$.

\section{Implementation details}
\label{supp:impl_details}

\subsection{RGB head: architecture and training}
\label{supp:rgb_head}

Since VGGT~\cite{wang2025vggt} produces depth, point, and camera predictions but no RGB output, we train a dedicated RGB head on frozen VGGT tokens. Like VGGT's native prediction heads, our RGB head consumes the concatenated patch tokens from the same encoder blocks $4,11,17,23$, $\token = \bigoplus_{i \in \{4, 11, 17, 23\}} h_i$, with $\token \in \R^{4 \cdot 2 \cdot 1024}$. The RGB head reconstructs a 3-channel RGB image at the input resolution of the images used to obtain the latent tokens. Its design follows a masked-autoencoder (MAE) ViT decoder trained with
$\ell_1$ + LPIPS + a patch-GAN, following the general design of GLD~\cite{jang2026repurposing}.

\paragraph{Architecture}
The decoder is a non-hierarchical ViT that operates on each view independently, meaning that there is no cross-view attention. Given patch tokens $\mathbf{X}\in\mathbb{R}^{N\times 8192}$:
\begin{enumerate}\itemsep0pt
  \item a shared \texttt{LayerNorm}$(2048)$ is applied independently to each of the four feature slices, matching the normalization used by VGGT's DPT heads,
  \item the concatenated $8192$-dimensional token is projected to decoder width $D$,
  \item a learned \textsc{cls} token is prepended and a fixed 2D sine-cosine positional embedding, 
  \item $L$ pre-norm transformer blocks with multi-head self-attention and GELU MLPs are applied,
  \item a final \texttt{LayerNorm} and linear projection $D\to14\cdot14\cdot3$ predict RGB pixels for each patch, which are subsequently unpatchified.
\end{enumerate}
The decoder predicts ImageNet-normalized RGB, which is de-normalized to
$[0,1]$ for evaluation. Architectural details are summarized in~\cref{tab:mae_gan} (right).

\begin{table}[t]
  \centering
  \caption{\textbf{RGB decoder.} Left: reconstruction from GT VGGT tokens on ScanNet++ validation and RE10K test views. Right: decoder
architecture. The parameter count includes the $8192\to D$ input projection and excludes the fixed positional embedding.}
  \label{tab:mae_gan}

  \begin{minipage}[t]{0.56\linewidth}
    \centering
    \setlength{\tabcolsep}{5pt}
    \begin{tabular}{l ccc}
      \toprule
      Dataset & PSNR\,$\uparrow$ & SSIM\,$\uparrow$ & LPIPS\,$\downarrow$ \\
      \midrule
      ScanNet++      & 27.63 & 0.897 & 0.090 \\
      RealEstate10K  & 25.68 & 0.838 & 0.114 \\
      \bottomrule
    \end{tabular}
  \end{minipage}\hfill
  \begin{minipage}[t]{0.40\linewidth}
    \centering
    \setlength{\tabcolsep}{5pt}
    \begin{tabular}{l r}
      \toprule
      Component & Value \\
      \midrule
      Width $D$   & 768 \\
      Layers $L$  & 12 \\
      Heads       & 12 \\
      MLP         & 3072 \\
      Params      & 91.8M \\
      \bottomrule
    \end{tabular}
  \end{minipage}
\end{table}

\paragraph{Training data}
The RGB head is trained on the ScanNet++ and RealEstate10K training splits, sampled with equal probability. Sequence lengths are drawn from
${2,3,4,5}$. Since decoding is performed independently per view, a batch of $B$ sequences containing $S$ views contributes $B\cdot S$ independent image
reconstructions. We use $B=8$ sequences per GPU, corresponding to $16$-$40$ images per GPU and step ($28$ on average).

\paragraph{Objective}
Let $x_{\mathrm{rgb}}$ denote a ground-truth frame and $\hat{x}_{\mathrm{rgb}}$ its reconstruction obtained by encoding the frame
with frozen VGGT and decoding the resulting tokens. The generator objective is

\begin{equation}
\mathcal{L}_G \;=\; \underbrace{\lVert \hat{x}_{\mathrm{rgb}}-x_{\mathrm{rgb}} \rVert_1 \;+\; \mathrm{LPIPS}(\hat{x}_{\mathrm{rgb}},x_{\mathrm{rgb}})}_{\textstyle \mathcal{L}_{\mathrm{rec}}}
\;+\; 0.75\,\lambda_{\mathrm{ada}}\;\underbrace{\bigl(-\mathbb{E}\,D(\mathcal{A}(\mathcal{C}(\hat{x}_{\mathrm{rgb}})))\bigr)}_{\textstyle \mathcal{L}_{\mathrm{gan}}},
\label{eq:maegan_g}
\end{equation}

where LPIPS uses the VGG-16 variant of~\cite{lpips}, evaluated on inputs mapped
to $[-1,1]$, $\mathcal{C}$ denotes a random $224\times224$ crop,
$\mathcal{A}$ differentiable augmentation, and $D$ the discriminator described
below. The $\ell_1$ term is the mean absolute error over pixels and color
channels. Both reconstruction terms have unit weight and are active from the
start of training.

Following VQGAN~\cite{esser2021taming}, the adversarial term is rescaled by the adaptive weight
\begin{equation}
\lambda_{\mathrm{ada}} \;=\;
\mathrm{clip}\!\left(
\frac{\lVert \nabla_{W}\mathcal{L}_{\mathrm{rec}}\rVert_2}
{\lVert \nabla_{W}\mathcal{L}_{\mathrm{gan}}\rVert_2+10^{-6}},
\;0,\;10^{4}
\right),
\label{eq:maegan_ada}
\end{equation}
computed with respect to the weights $W$ of the final prediction layer.

The discriminator is trained with the hinge loss
\begin{equation}
\mathcal{L}_D \;=\; \tfrac{1}{2}\Bigl( \mathbb{E}\,\max\!\left(0,1-D(\mathcal{A}(\mathcal{C}(x_{\mathrm{rgb}})))\right)
+
\mathbb{E}\,\max\!\left(0,1+D(\mathcal{A}(\mathcal{C}(\hat{x}_{\mathrm{rgb}})))\right) \Bigr),
\label{eq:maegan_d}
\end{equation}
and returns one logit per patch position rather than a single score per image. $\mathbb{E}$ denotes a uniform mean over all discriminator
outputs and all images in the batch. Within each discriminator step, the same crop window is applied to $x_{\mathrm{rgb}}$ and $\hat{x}_{\mathrm{rgb}}$.

Differentiable augmentation follows StyleGAN-T~\cite{sauer2023stylegan} and
consists of random translation by up to $\pm12.5\%$ together with
brightness, saturation, and contrast jitter, applied with probability $1$.
Augmentation is used in both generator and discriminator updates. The
discriminator is trained from step 8k onward, and the generator
receives the adversarial term from step 12k onward; before that,
training is purely reconstructive.

\paragraph{Discriminator}
We use a frozen DINO ViT-S/8~\cite{caron2021emerging} patch discriminator with features extracted after blocks $\{2,5,8,11\}$. Following
GLD~\cite{jang2026repurposing}, five lightweight heads operate on these features, with two heads applied to the final block. Each head is a lightweight convolutional network with spectral normalization and LeakyReLU activations, producing per-token discriminator logits used in
\cref{eq:maegan_g,eq:maegan_d}. Only the discriminator heads are trained ($7.4$M parameters).

\paragraph{Optimization}
We use Adam~\cite{kingma2014adam} with $\beta=(0.9,0.95)$ and no gradient clipping. The decoder uses a learning rate of $2\times10^{-4}$ with 5k
warm-up steps, followed by cosine decay to $2\times10^{-5}$ at step
$100$k; the discriminator uses the same learning-rate range with
$2$k warm-up steps and decay completed at step $60$k. We maintain an
EMA of the decoder weights with decay $0.9978$ and use the EMA model for
evaluation. Training runs for $200$k steps on four H100 GPUs, corresponding
to approximately $22.4$M decoded images and $53$ hours.

\paragraph{Reconstruction quality}
\Cref{tab:mae_gan} reports reconstruction from ground-truth VGGT tokens, i.e.
real images encoded by the frozen backbone and decoded by the RGB head. These results characterize the fidelity of the decoder independently of the
generative model and provide a reference for the RGB quality attainable from
real VGGT tokens. They should not be interpreted as novel view generation results.

\subsection{Training details of our Riemannian flow matching model}
\label{app:training_details}

We train our generative model for $330$k steps using Adam~\cite{kingma2014adam} without weight decay. The learning rate is linearly warmed up for the first
$10$k steps to $10^{-3}$ and then cosine-decayed to $10^{-6}$. We maintain an exponential moving average (EMA) of the model parameters with decay $0.9999$
and use the EMA weights at inference. Gradients are clipped to a maximum norm of $1.0$. Training uses bfloat16 mixed precision and a global batch size of
$256$ frame sequences distributed across $16$ H100 GPUs. Training takes approximately six days, dominated by the VGGT forward passes, which are computed online for every batch rather than precomputed.

Frame sequences are sampled with equal probability from RealEstate10K and
ScanNet++. Within each scene, we randomly sample the target view, context
views, and the number of context views $N\in\{1,2,3,4\}$, exposing the model
during training to varying conditioning configurations.

Rather than sampling flow times uniformly, $t\sim\mathcal{U}(0,1)$, we use a
shifted uniform distribution that places more probability mass near the noisy
end of the trajectory. We first sample
$\tau \sim \mathcal{U}(0,1)$
and transform it as $t  =  \frac{\tau}{\tau+s(1-\tau)}$.
Under our convention, $t=0$ corresponds to noise and $t=1$ to clean tokens, therefore $s>1$ biases training toward noisier states. We use $s=6$ in all
experiments.

\subsection{Baselines}
\label{app:baselines}
The compared methods differ in their conditioning interfaces and assumptions.
Our model accepts a variable number of unposed context images together with an
arbitrary target camera pose and does not require a prescribed camera
trajectory. In contrast, Gen3R operates along a camera trajectory, DepthSplat
requires at least two posed context views, and Matrix3D and GLD also use known
context poses. We therefore evaluate each baseline using its native interface
while matching the context and target views as closely as possible. Below, we
detail the checkpoints, inputs, and inference settings used for each method.

\paragraph{Gen3R}
Gen3R~\cite{huang2026gen3r} is a video-diffusion novel-view
synthesizer built on the VGGT geometry backbone. Unlike our method, it expects a
full camera trajectory and iteratively extends a partially observed scene, rather
than being conditioned on a single arbitrary target pose. To ensure a fair
comparison, for each context-target pairing we condition on the context view(s)
and synthesize the target by generating a short video along a camera trajectory
that passes through the target pose. The novel view is then read off at the trajectory index assigned to the target camera. We follow Gen3R's official preprocessing throughout.

\emph{Trajectory.} We use a $49$-frame anchor-based trajectory, with one
generation pass per target. The anchors are the context and target cameras: each
anchor's pose is kept exactly at its assigned index, and intermediate frames
between two consecutive anchors are filled by interpolation. Rotations are interpolated using spherical linear interpolation (SLERP), and translations linearly. 

\emph{One-sided vs.\ two-sided.} In the two-sided setting the
context views are spread evenly around the target, so the target is an interpolation between them
(context\,$\rightarrow$\,target\,$\rightarrow$\,context). In the one-sided setting the context
views are the ones farthest from the target, all lying on one side, so the target is reached by
extrapolating the camera path outward. 

\emph{Sampling.} We use classifier-free guidance with scale $5.0$, the text
prompt \texttt{"a realistic scene"} and the negative prompt \texttt{"bad detailed"} during sampling.

\paragraph{DepthSplat}
DepthSplat~\cite{xu2025depthsplat} is a deterministic feed-forward method that combines
multi-view stereo with 3D Gaussian splatting, predicting depth and Gaussians in a
single forward pass. We use the public \texttt{gs-large} checkpoint
(\texttt{depthsplat-gs-large-re10k-256x256-view2}), trained on RealEstate10K at
$256\times256$. The encoder is a large ViT with a cost-volume module whose depth candidates span
$[\text{near}, \text{far}]$, sampled uniformly in inverse depth. We set this range per
dataset to match each domain's metric extent: $0.5$--$100$ for RealEstate10K
(the training values) and $0.1$--$10$ for ScanNet++ (estimated from ScanNet++ training scenes). Target depth is obtained as the z-buffer of the splatted Gaussians at the target pose, so RGB and depth share the same $256\times256$ resolution.

\paragraph{Matrix3D} 
Matrix3D~\cite{lu2025matrix3d} is a latent multi-modal diffusion transformer that jointly models RGB, camera (Plücker ray maps), and depth under masked any-to-any conditioning, where each (view, modality) cell is marked condition, generate, or unused. Because target poses can be supplied as conditions, it addresses our task directly. We use the public \texttt{matrix3d\_512.pt} checkpoint in all experiments. Following the authors' instructions, target RGB and depthmaps are generated in two passes: target RGB at guidance 1.5 (50 DDPM steps), then depth at guidance 1.0, conditioned on the generated RGB. Pointmaps are obtained by back-projecting the predicted depth given no native pointmap head.

Matrix3D predicts depth in normalized scene units, and fixes that unit from the input poses alone if no ground truth depth is provided. It sets the largest camera-to-first-camera distance in the sequence to 1.0.

If necessary, when this derived unit falls outside its training distribution, we set the scale ourselves. To do so we rescale each sequence poses before inference so the expected scene depth is $\approx1$, by $k\times \text{spread}$, where spread is the maximum pairwise camera distance and $k$ is a single global constant per dataset directly calculated by the evaluation scenes (ScanNet++ 2.296, ETH3D 1.431), and undo the factor on the outputs.

\paragraph{GLD}
GLD~\cite{jang2026repurposing} performs multi-view generation in the feature space of a frozen geometric foundation
model. We use the VGGT variant with the public checkpoints for their best configuration, the \emph{cascade} mode
(\texttt{vggt\_level1.pt} and \texttt{vggt\_cascade.pt} ). The latent space consists of VGGT patch tokens from encoder blocks $[4, 11, 17, 23]$ and the cascade first diffuses the level-1 (block-11) tokens and then maps them to level 0, with levels 2/3 obtained by propagation. Sampling uses a linear rectified-flow ODE (Euler, 100 steps),
classifier-free guidance with scale $1.5$. RGB is decoded by its own RGB head, depth maps by the frozen VGGT depth head, and point maps from the frozen VGGT pointmap head, all from the same synthesized tokens.

\section{Additional details of the evaluation protocol}
\label{app:eval_protocol}

\subsection{Evaluation scenes}
\label{app:eval_scenes}
All novel-view results are computed on $50$ scenes of the ScanNet++ validation split and on $50$ scenes of the RE10K test split, the latter chosen as scenes in which context--target windows satisfying the overlap criteria of \cref{supp_sub:overlap_cats} exist. For 3D scan generation, ScanNet++ uses the same $50$ scenes and ETH3D averages are over all $13$ scenes of its training split.

\subsection{Resolution unification for cross-method evaluation}
\label{app:unification}

Each method we compare against preprocesses its inputs differently and therefore
predicts a different sub-region of the scene at a different resolution. Per-pixel metrics such as PSNR, SSIM, RMSE, ARel and $\delta$ are only comparable if every method is scored on the same scene region sampled at the same pixel grid. Naively scoring each method on its own preprocessed output would conflate synthesis quality with differences in field of view (FOV) and sampling density.

We resolve this with a single common grid per dataset, constructed so that all methods are evaluated on exactly the scene region that all of them predict. Both the predictions and the
ground truth are mapped onto this grid before any metric is computed, including
the distributional FID.

\paragraph{Common-grid construction}
Let the original frame have resolution $(H, W)$. For each method $m$, we determine the rectangle of the original frame it actually predicts, expressed in original pixel coordinates, which we call its FOV $\mathcal{F}_m = (H_m^{\mathrm{fov}},
W_m^{\mathrm{fov}})$. The FOVs are obtained by inverting each method's preprocessing:

\begin{itemize}
    \item \textbf{DepthSplat} fits the original into its $256\times256$ crop with
        the cover scale $\sigma = \max(256/H,\, 256/W)$, retaining
        $\mathcal{F}_{\text{D}} = \bigl(\min(\lfloor 256/\sigma \rceil, H),\,
        \min(\lfloor 256/\sigma \rceil, W)\bigr)$.
    \item \textbf{Gen3R} resizes the shorter side to $560$ and center-crops a
        square, so it predicts a centered square of the original:
        $\mathcal{F}_{\text{G}} = (s, s)$ with $s = \min(H, W)$.
    \item \textbf{Matrix3D} centre-crops the original to the largest centred
        square and resizes it to $512\times512$, so
        $\mathcal{F}_{\text{M}} = (s, s)$ with $s = \min(H, W)$ (its depth is
        predicted at a native $128\times128$; see below).

    \item \textbf{GLD} resizes the shorter side to $504$ and center-crops a
        square, giving the same centered square
        $\mathcal{F}_{\text{GLD}} = (s, s)$, $s = \min(H, W)$.

    \item \textbf{Ours} resizes width to $518$, sets height to the nearest
        multiple of $14$, and center-crops/pads to the canonical
        $294\times518$. This keeps the full original width and crops height, so
        $\mathcal{F}_{\text{O}}$ retains the full width and a centred vertical
        band of the original.
\end{itemize}

The only region predicted by \emph{all} methods is the elementwise intersection of these rectangles:
\begin{equation}
  \mathcal{F}_{\cap}
  = \Bigl(\min_m H_m^{\mathrm{fov}},\ \min_m W_m^{\mathrm{fov}}\Bigr).
\end{equation}

To avoid upsampling any method, we sample $\mathcal{F}_{\cap}$ at the coarsest native resolution among the methods (the one exception is Matrix3D's depth, whose native $128\times128$ lies below this density and is nearest-neighbour upsampled to the grid; its per-pixel depth metrics are unchanged when re-scored at its native rate). For method $m$ with output size $S_m$ on a given axis covering an FOV extent $F_m$ of the original, the native sampling density is $\rho_m = S_m / F_m$ (output pixels per original pixel). 
We use a single isotropic scale
\begin{equation}
  r = \min_m \rho_m,
\end{equation}
taken over both axes of all methods, which preserves the aspect ratio of
$\mathcal{F}_{\cap}$ and guarantees $r \le \rho_m$ for every method, i.e. no method is upsampled. The common grid is
\begin{equation}
  (H_g, W_g) = \bigl(\lfloor r\, H^{\mathrm{fov}}_{\cap} \rceil,\
                       \lfloor r\, W^{\mathrm{fov}}_{\cap} \rceil\bigr).
\end{equation}

Applying this construction yields a fixed grid per dataset, summarised in~\cref{tab:common-grids} and an illustration for the ScanNet++ dataset is given in~\cref{fig:fov}
.

\begin{figure}[t]
  \centering
  \begin{minipage}[c]{0.55\linewidth}
    \centering
    \resizebox{\linewidth}{!}{%
    \begin{tikzpicture}
      \fill[green!55!black, opacity=0.22] (-2.336,-1.962) rectangle (2.336,1.962);
      \draw[gray!70, line width=0.6pt, dashed] (-3.504,-2.336) rectangle (3.504,2.336);
      \node[gray!55!black, font=\scriptsize, anchor=north west] at (-3.504,-2.40)
        {Original frame $1752\times1168$};
      \draw[orange!85!black, line width=1.1pt] (-3.504,-1.962) rectangle (3.504,1.962);
      \node[orange!70!black, font=\scriptsize\bfseries, anchor=west] at (3.55,1.50) {Ours};
      \node[orange!70!black, font=\scriptsize, anchor=west] at (3.55,1.18) {(binds H)};
      \draw[blue!75!black, line width=1.1pt] (-2.336,-2.336) rectangle (2.336,2.336);
      \node[blue!70!black, font=\scriptsize\bfseries, anchor=south] at (0,2.52)
        {Gen3R / DepthSplat / GLD / Matrix3D};
      \node[blue!70!black, font=\scriptsize, anchor=south] at (0,2.22) {(bind W)};
      \draw[green!45!black, line width=1.3pt] (-2.336,-1.962) rectangle (2.336,1.962);
      \node[green!35!black, font=\scriptsize\bfseries, align=center] at (0,0)
        {Intersection FOV\\ $981\times1168$\\[2pt]
         \footnotesize $\to$ common grid $215\times256$};
    \end{tikzpicture}%
    }
    \caption{\textbf{Resolution unification on ScanNet++.} Gen3R, DepthSplat, GLD and
      Matrix3D bind the width; our method binds the height. Their green
      intersection defines the $215\times256$ common grid. The four square-crop
      methods coincide on these datasets; in general they need not.}
    \label{fig:fov}
  \end{minipage}\hfill
  \begin{minipage}[c]{0.42\linewidth}
    \centering
    \captionof{table}{\textbf{Common evaluation grids.} Grids resulting from the unification, fixed per dataset. ETH3D DSLR images vary slightly across scenes, hence the $\approx$.}
    \label{tab:common-grids}
    \setlength{\tabcolsep}{3pt}
    \footnotesize
    \begin{tabular}{lcc}
  \toprule
  Dataset & Orig.\ ($W\!\times\!H$) & Grid ($H\!\times\!W$) \\
  \midrule
  RE10K     & $640\times360$          & $256\times256$ \\
  ScanNet++ & $1752\times1168$        & $215\times256$ \\
  ETH3D     & ${\approx}\,6205\times4135$ & $215\times256$ \\
  \bottomrule
\end{tabular}
  \end{minipage}
\end{figure}

\paragraph{Mapping predictions and ground truth onto the grid}
Each method's prediction is center-cropped to the intersection sub-rectangle within its own output frame and then resampled to $(H_g, W_g)$. The ground truth is center-cropped from the original frame to $\mathcal{F}_{\cap}$ and
resampled identically. For RGB, we resample with Lanczos interpolation. For depth,
we resample with nearest-neighbor. Invalid pixels are preserved
exactly rather than averaged into their neighbors. All metrics are then computed on the common grid.

\subsection{3D metric scales alignment}
\label{app:metricalignment}

VGGT's predicted depth and point maps are up to an unknown scale and shift. So before calculating the metrics, we align predictions to the ground truth's metric scale. The same procedure is applied to every method.

\paragraph{Depth maps} For depth maps, we do so by a single least-squares scale-and-shift fit over the valid pixels of the first context frame:
\begin{equation}
  (a^\star, b^\star) = \argmin_{a, b}
  \sum_{i \in \mathrm{valid}} \bigl(a\, \hat{d}_i + b - d_i\bigr)^2,
\end{equation}
where $d_i$ and $\hat{d}_i$ are the ground truth and predicted depth for a valid pixel $i$.
The valid set of pixels consists of those for which the ground truth depth is positive and the prediction is finite and positive. Pixels without a sensor return (or without a laser hit) are excluded from the fit. The resulting affine is
then applied to the predicted target frame,
\begin{equation}
  \tilde{d}_i = a^\star\, \hat{d}_i + b^\star.
\end{equation}
If fewer than $1{,}000$ valid pixels support the fit, the sample falls back to fitting on the target frame itself. This is applied identically to all methods.
Depth metrics are calculated over all valid pixels after alignment.

\paragraph{Pseudo-depth evaluation on RE10K} RE10K provides no ground-truth depth, so we score depth predictions against a pseudo ground truth from VGGT. For every context target pairing, target view and number of context views $N$, we run VGGT once on the $N$ context images together with the ground-truth target image and take its predicted depth of the target view as the reference. All methods are scored against the same VGGT geometry. Predictions and pseudo ground truth are mapped to the common $256\times256$ grid (see~\cref{app:unification}) and compared on valid pixels without any scale or shift alignment: since all compared methods produce depth in VGGT's normalised scale (the per-sample median scale ratio to the reference is 0.99–1.03 for every method).

\paragraph{Point maps} For point maps we align via a similarity transform $(R, s, t) \in SO(3) \times \mathbb{R}_{>0} \times \mathbb{R}^3$ using Umeyama's closed-form solution~\cite{umeyama1991},
\begin{equation}
(R^\star, s^\star, t^\star) = \argmin_{R, s, t}
\sum_{i \in V_{\mathrm{ctx}}} \bigl\lVert s\, R\, \hat{p}_i + t - p_i \bigr\rVert^2,
\end{equation}
where $\hat{p}_i$ and $p_i$ are corresponding predicted and ground-truth world-space points. We write $V_F$ for the joint validity mask over a set of frames $F$: those pixels at which the predicted point is finite and the ground-truth depth is positive. The fit uses $V_{\mathrm{ctx}}$, the context frames only, and the single resulting transform is applied uniformly to every frame's point cloud.

To prevent a small number of noisy predicted points from biasing the fit we apply Umeyama in two passes. A tentative fit brings the predicted cloud into the ground-truth frame. We then run a statistical outlier filter on the aligned prediction, dropping any point whose mean distance to its $25$ nearest neighbors exceeds $\mu + 1.25\,\sigma$ (with $\mu$ and $\sigma$ the mean and standard deviation of these per-point mean distances across the cloud). $V_{\mathrm{ctx}}$ is updated to exclude the removed points, and Umeyama is re-fit on the cleaned correspondences.

Each method's point maps are obtained as follows. Our method generates each target in its own pass from the real context tokens and decodes contexts and both targets through VGGT's point head once, giving one cloud and one fit. GLD, Gen3R and Matrix3D predict both targets jointly in one pass, so their clouds are likewise aligned once on the context frames. Matrix3D has no point-map head, so its point maps are obtained by back-projecting its depths of all four views with the same given cameras that conditioned the generation. Ground-truth point maps are back-projected from the original-resolution depth, predicted and ground-truth maps are mapped onto the common grid of~\cref{app:unification} before correspondences are formed.

Metrics are computed over the context and generated target frames jointly ($V_{\mathrm{ctx}\cup\mathrm{tgt}}$). The predicted and ground-truth clouds are restricted to this joint mask, concatenated across frames, and then uniformly voxel-downsampled (voxel size $5$~mm) followed by farthest-point subsampling to at most $K = 20$k points per side (skipped when a cloud already has fewer points), so that scenes are compared on comparable point budgets rather than raw per-frame pixel counts. No outlier filtering is applied to the merged predicted cloud before scoring. We report accuracy (mean nearest-neighbour distance from prediction to ground truth), completeness (ground truth to prediction), their mean (Chamfer), and Chamfer relative to the ground-truth bounding-box diagonal.

\subsection{Reconstructed and generated region split}
\label{app:rec_gen}

To separate \emph{reconstruction} of observed content from \emph{hallucination}
of occluded or unseen content, we additionally split each unified target view into a
visible and a generated region. The split is a per-pixel mask over the same
unified pixels (see~\cref{app:unification}). Crucially, the mask depends only on ground-truth geometry and is thus identical across methods, making the per-region numbers directly comparable.

A target pixel is labeled \textit{visible} if back-projecting its ground-truth
depth into world space and reprojecting it into at least one context view lands
on a consistent depth, $|z_{\text{ctx}} - D_{\text{ctx}}(\pi)| < \tau\,
D_{\text{ctx}}$ with $\tau = 0.05$ (relative). It is labeled \textit{generated} if the target
pixel has valid depth but is seen by no context view. The same mask is shared across RGB and depth, since both
live on the same grid.

On the visible region, we report reconstruction metrics (PSNR/SSIM/LPIPS for RGB,
RMSE/ARel/$\delta$ for depth). On the generated region we report
the generation metric FID. For depth, we apply a single global scale-and-shift
alignment over all valid pixels as explained in~\cref{app:metricalignment} and only then split.

\paragraph{Region-wise LPIPS and FID}
LPIPS was computed on the full unified $215\times256$ image with non-visible pixels zeroed in both prediction and ground truth before passing to AlexNet-LPIPS, the identical masking on both sides drives the non-visible contribution toward zero, so the reported value is dominated by the visible pixels.
FID was accumulated over the generated region only via a composite-onto-GT variant, $\mathrm{fake} = \mathrm{gt} \cdot (1 - m_{\mathrm{gen}}) + \mathrm{pred} \cdot m_{\mathrm{gen}}$ against $\mathrm{real} = \mathrm{gt}$, so the distribution shift measured by FID isolates the hallucinated pixels.

\subsection{High and low overlapping categories description}
\label{supp_sub:overlap_cats}

In order to evaluate a smooth transition from reconstruction to generation, the model's ability to generate consistent and plausible geometry for seen as well as hallucinated areas, we categorize context and target pairings in \emph{high overlapping} or
\emph{low overlapping} categories, based on the relative camera poses. 

Pairings are characterized by three measures computed from the world-to-camera extrinsics:

\begin{itemize}
    \item \textbf{Rotation} $\theta_{\text{rot}}$ (degrees): the geodesic distance (angle) in SO(3)
          between the two rotation matrices.
    \item \textbf{Lateral translation} $t_{\text{lat}}$ (meters): the magnitude
          of the relative camera position projected onto the context camera's
          image plane (i.e.\ the $x$-$y$ components in context-camera coordinates).
    \item \textbf{Longitudinal translation} $t_{\text{long}}$ (meters): the
          absolute value of the context-camera $z$-component of the relative
          position (motion along the optical axis).
\end{itemize}

The rotation threshold is \emph{scaled by the context camera's horizontal
field of view} $\mathrm{FoV}_x$, since a fixed angular displacement removes a
larger fraction of pixels under a narrow lens. Translations are absolute.
A context - target pairing is labeled \emph{high overlapping} if there is one context frame for the target frame at hand, where all three measures lie within
the high-overlap bounds $(\theta^{\text{h}}, \ell^{\text{h}}, z^{\text{h}})$,
and \emph{low overlapping} if all three measures lie within the looser
low-overlap bounds $(\theta^{\text{l}}, \ell^{\text{l}}, z^{\text{l}})$ but
the pairing is not already classified as high overlapping. Table~\ref{tab:thresh}
lists the FoV-relative bounds. We compute $\mathrm{FoV}_x$ from the
camera intrinsics of the ground truth, which gives
$\mathrm{FoV}_x \approx 109^\circ$ on ScanNet++ and
$\mathrm{FoV}_x \approx 91^\circ$ on RE10K; the absolute rotation bounds are
then $\theta^{\text{h}}_{\text{ScanNet++}} \approx 11^\circ$,
$\theta^{\text{l}}_{\text{ScanNet++}} \approx 66^\circ$,
$\theta^{\text{h}}_{\text{RE10K}} \approx 14^\circ$,
$\theta^{\text{l}}_{\text{RE10K}} \approx 64^\circ$.

\begin{table}[t]
\scriptsize
    \centering
    \caption{\textbf{Overlap-tier thresholds.} Thresholds used to classify context--target view pairings. Rotation bounds are
             relative to the context camera's horizontal field of view
             ($\mathrm{FoV}_x \approx 109^\circ$ on ScanNet++,
             $\approx 91^\circ$ on RE10K); translations are absolute (m).}
    \label{tab:thresh}
    \begin{tabular}{l ccc ccc}
        \toprule
        & \multicolumn{3}{c}{ScanNet++} & \multicolumn{3}{c}{RE10K} \\
        \cmidrule(lr){2-4} \cmidrule(lr){5-7}
        Category
            & $\theta_{\text{rot}}$          & $t_{\text{lat}}$ & $t_{\text{long}}$
            & $\theta_{\text{rot}}$          & $t_{\text{lat}}$ & $t_{\text{long}}$ \\
        \midrule
        High overlap
            & $\leq 0.10\,\mathrm{FoV}_x$ & $\leq 0.20$ & $\leq 0.50$
            & $\leq 0.15\,\mathrm{FoV}_x$ & $\leq 0.25$ & $\leq 0.70$ \\
        Low overlap
            & $\leq 0.60\,\mathrm{FoV}_x$ & $\leq 1.50$ & $\leq 3.00$
            & $\leq 0.70\,\mathrm{FoV}_x$ & $\leq 1.50$ & $\leq 3.20$ \\
        \bottomrule
    \end{tabular}
\end{table}

RE10K's bounds are widened relative to ScanNet++ to account for the smoother, slower camera motion typical of real-estate captures. To prevent high-overlap 2-sided sequences from collapsing onto temporally adjacent, meaning almost identical, frames on RE10K, every edge of the four-context setup ($S_0 \to S_1 \to T \to S_2 \to S_3$) and every context-target pair is required to span at least 12 frames. ScanNet++ uses a frame gap of 1, since the aequidistant subsampling to 150 frames per scene already provides a wider gap.
\input{imgs/protocols}
An illustration of the one-sided vs. two-sided evaluation setup is given in~\cref{fig:sequence_layout}.

\paragraph{Percentage overlap} We measure the actual per-pixel overlap between each
target view and its context views on the unified evaluation grid
(\cref{tab:overlap}). For ScanNet++ we compute overlap exactly from
ground-truth depth; for RE10K, which lacks GT depth, we use a constant-depth
($Z\!=\!4$\,m) proxy, so the two datasets are not directly comparable in
absolute terms. We report the mean fraction of target pixels covered under four
context-aggregation rules, the union of all context frames, and the maximum, median,
and minimum coverage from any single context frame, which together bracket the
best- and worst-case visibility a model can exploit.

On ScanNet++, coverage drops sharply from the high- to the low-overlap tier under every aggregation
rule (e.g.\ union coverage falls from $0.82$ to $0.47$ one-sided and from $0.92$
to $0.79$ two-sided), so the low-overlap tier genuinely demands more generation
of unobserved content. The two-sided setup consistently yields higher coverage
than the one-sided one, as bracketing the target between context views leaves
fewer unseen regions. On RE10K the overlaps are uniformly higher even in the
low-overlap tier ($\geq 0.89$ union), reflecting the smooth, forward-facing
motion of real-estate captures.

For the ETH3D scenes used in the 3D scan generation
experiment, we report the analogous overlap statistics in
\cref{tab:eth3d-legacy-pixel-overlap}. Here, the overlap is again computed exactly
from ground-truth depth on the unified $215\times256$ crop, over all
$n\!=\!26$ target samples ($13$ scenes, one two-context/two-target window each),
and we report the median with $[\text{p5},\text{p95}]$ range rather than the
mean, given the small sample count. Even under the union of both context frames the
median target-pixel coverage is only $0.61$, and drops to $0.29$ for the
worst single context frame. This makes ETH3D a substantially harder and sparser setting
than ScanNet++.

\begin{table}[t]
  \centering
  \begin{minipage}[t]{0.55\linewidth}
    \centering
    \footnotesize
    \setlength{\tabcolsep}{8pt}
    \caption{\textbf{Context--target overlap on ScanNet++ and RE10K.} Mean fraction of
      target pixels covered by the context views on the unified evaluation grid, under
      four aggregation rules (union of all context views; max / median / min over single
      context views). ScanNet++: exact overlap from GT depth; RE10K: constant-depth
      ($Z\!=\!4$\,m) proxy.}
    \label{tab:overlap}
    \begin{tabular}{c l cc cc}
      \toprule
      & & \multicolumn{2}{c}{ScanNet++} & \multicolumn{2}{c}{RE10K} \\
      \cmidrule(lr){3-4}\cmidrule(lr){5-6}
      & & high & low & high  & low \\
      \midrule
      \multirow{5}{*}{\rotatebox[origin=c]{90}{one-sided}}
        & $n_\text{targets}$ & 500  & 500  & 500  & 500  \\
        & union-of-4       & 0.82 & 0.47 & 0.93 & 0.89 \\
        & max-ctx          & 0.78 & 0.44 & 0.93 & 0.89 \\
        & med-ctx          & 0.59 & 0.31 & 0.93 & 0.88 \\
        & min-ctx          & 0.42 & 0.21 & 0.92 & 0.87 \\
      \midrule
      \multirow{5}{*}{\rotatebox[origin=c]{90}{two-sided}}
        & $n_\text{targets}$ & 500  & 494  & 364  & 228  \\
        & union-of-4       & 0.92 & 0.79 & 1.00 & 1.00 \\
        & max-ctx          & 0.85 & 0.60 & 1.00 & 1.00 \\
        & med-ctx          & 0.65 & 0.38 & 0.95 & 0.90 \\
        & min-ctx          & 0.44 & 0.17 & 0.82 & 0.59 \\
      \bottomrule
    \end{tabular}
  \end{minipage}
  \hfill
  \begin{minipage}[t]{0.42\linewidth}
    \centering
    \footnotesize
    \setlength{\tabcolsep}{10pt}
    \caption{\textbf{Context--target overlap on ETH3D.} Same quantity as
      \cref{tab:overlap}, computed exactly from GT depth over valid target pixels;
      median across the $26$ samples with $[\text{p5},\text{p95}]$ in brackets.
      Union-of-2 counts pixels visible from either context view; max / med / min-ctx
      report the best / typical / worst single-context-view coverage.}
    \label{tab:eth3d-legacy-pixel-overlap}
    \footnotesize
    \begin{tabular}{@{}lc@{}}
      \toprule
      \multicolumn{2}{c}{ETH3D ($n_\text{targets} = 26$)}\\
      \midrule
      & Median [p5, p95] \\
      \midrule
      Union-of-2 & 0.61 [0.47, 0.89] \\
      Max-ctx    & 0.45 [0.29, 0.83] \\
      Med-ctx    & 0.38 [0.27, 0.64] \\
      Min-ctx    & 0.29 [0.01, 0.59] \\
      \bottomrule
    \end{tabular}
  \end{minipage}
\end{table}

\section{Additional experiments and ablations}
\label{supp:add_ablations}

Unless stated otherwise, all ScanNet++ results in this section follow the evaluation protocol of the main paper: pixel-aligned metrics (PSNR, SSIM, LPIPS, RMSE, ARel, $\delta$) are computed on the regions observed in the context views and FID on the unobserved regions (\cref{app:rec_gen}).

\subsection{Context order shuffling}

Our formulation does not assume a fixed ordering or camera trajectory for the context views. To verify this, we evaluate the same $N{=}4$ context set under different permutations in \cref{tab:context_order_full}. Across three random orderings, performance is highly stable: in the high-overlap setting, the standard deviation is only $0.01$\,dB in PSNR and $0.003$ in depth RMSE, with
similarly small variation across the remaining metrics. The low-overlap regime is equally stable. Thus, unlike trajectory-based generation methods, our model does not rely on a particular temporal or spatial ordering of the context views.

A small systematic effect appears when the context view closest to the target
is placed first. This ordering yields slightly better performance, most
noticeably in the high-overlap setting (e.g.\ PSNR $20.55$ vs.\ $20.01$ for
random orderings). We attribute this to the underlying VGGT representation,
for which the first input frame plays a distinguished role. Importantly, the small variance across random permutations shows that this positional bias does
not translate into a requirement for any fixed context trajectory or ordering.

\begin{table}[t]
  \centering
  \caption{\textbf{Robustness to context view ordering.} One-sided generation from $N{=}4$ context views on ScanNet++. Random \#1--\#3 are three random permutations of the context views.}
  \label{tab:context_order_full}
  \setlength{\tabcolsep}{5pt}
  \scriptsize
  \begin{tabular}{p{10pt}l rrrr rrrr}
    \toprule
    & & \multicolumn{4}{c}{RGB} & \multicolumn{4}{c}{Depth} \\
    \cmidrule(lr){3-6} \cmidrule(lr){7-10}
    & Context order & \multicolumn{1}{c}{PSNR\,$\uparrow$} & \multicolumn{1}{c}{SSIM\,$\uparrow$} & \multicolumn{1}{c}{LPIPS\,$\downarrow$} & \multicolumn{1}{c}{FID\,$\downarrow$}
                   & \multicolumn{1}{c}{RMSE\,$\downarrow$} & \multicolumn{1}{c}{ARel\,$\downarrow$} & \multicolumn{1}{c}{$\delta$\,$\uparrow$} & \multicolumn{1}{c}{FID\,$\downarrow$} \\
    \midrule
    \multirow{5}{*}{\rotatebox[origin=c]{90}{High overlap}}
    & Furthest first     & 19.62 & 0.689 & 0.111 & 26.19 & 0.218 & 0.103 & 0.905 & 24.80 \\
    & Closest first      & 20.55 & 0.722 & 0.091 & 24.69 & 0.202 & 0.091 & 0.916 & 24.35 \\
    & Random \#1    & 20.02 & 0.703 & 0.103 & 25.10 & 0.214 & 0.099 & 0.907 & 24.53 \\
    & Random \#2    & 20.02 & 0.703 & 0.103 & 25.37 & 0.207 & 0.095 & 0.913 & 24.38 \\
    & Random \#3    & 20.00 & 0.702 & 0.103 & 25.40 & 0.212 & 0.098 & 0.908 & 24.36 \\
    \midrule
    \multirow{5}{*}{\rotatebox[origin=c]{90}{Low overlap}}
    & Furthest first     & 17.65 & 0.628 & 0.110 & 51.82 & 0.287 & 0.120 & 0.871 & 43.95 \\
    & Closest first      & 17.85 & 0.635 & 0.107 & 52.03 & 0.284 & 0.119 & 0.874 & 43.66 \\
    & Random \#1    & 17.74 & 0.631 & 0.109 & 52.08 & 0.285 & 0.119 & 0.874 & 43.96 \\
    & Random \#2    & 17.75 & 0.632 & 0.108 & 51.92 & 0.285 & 0.119 & 0.872 & 43.82 \\
    & Random \#3    & 17.78 & 0.633 & 0.107 & 51.99 & 0.285 & 0.119 & 0.869 & 43.93 \\
    \bottomrule
  \end{tabular}
\end{table}

\subsection{Generality of the latent space}
\label{app:da3exp}

Our Riemannian flow-matching formulation is not specific to VGGT. It only requires a frozen backbone and decoder for which the features consumed by the decoding heads are LayerNormed. Given any set of such intermediate features, we apply LayerNorm and project the resulting features onto the unit sphere, and perform flow matching directly on the resulting manifold. If a single intermediate feature is used, the latent space is a single sphere, if multiple intermediate features are used, it is a product of spheres. The construction therefore does not depend on a particular backbone architecture, feature dimensionality, or number of intermediate features.

The geometric interpretation is also natural in high dimensions: high-dimensional representations already exhibit an approximate spherical structure, as their norms tend to concentrate near an outer shell. LayerNorm makes this structure explicit by normalizing each feature to a fixed scale, after which projection onto the unit sphere gives an explicit spherical manifold on which the flow is defined. The frozen decoder already applies the same normalization to the features it consumes, so generated representations can be passed to the decoder without architectural modification.

\paragraph{Application to Depth Anything 3} To test this generality, we apply the same formulation to DA3~\cite{lin2025depth}, another 3D backbone with a DINOv2 ViT architecture and alternating cross-view attention, QK normalization, and RoPE in its later blocks, trained for 3D reconstruction. For training in the DA3 latent space, we use the four intermediate features consumed by its DualDPT decoder (blocks $11,15,19,23$),
$\token = \bigoplus_{i \in {11,15,19,23}} h_i$, with $\token \in \R^{4 \cdot 2 \cdot 1024}$. Thus, the resulting latent dimensionality and product-manifold structure match those used for VGGT, allowing us to reuse the generative model architecture without modification.

Applying the same training recipe to DA3 produces a working model across all context settings and overlap regimes (Table~\ref{app_tab:da3_depth}). The DA3 model exhibits trends closely aligned with VGGT as the amount of context increases, with competitive depth accuracy and similar FID in both high- and low-overlap settings. While VGGT achieves somewhat stronger depth metrics overall, the comparable performance across the two backbones demonstrates that the proposed Riemannian flow-matching formulation transfers to a substantially different, independently pretrained 3D backbone without requiring changes to the generative formulation, latent geometry, or model architecture.

\begin{table}[t]
\vspace{-10pt}
  \centering
  \caption{\textbf{Transfer to a different 3D backbone.}
Novel view generation on ScanNet++ using our method with VGGT and DA3 latents. 
The same generative formulation and training recipe are used for both backbones. 
Results are reported for high- and low-overlap settings with one-sided and two-sided context.}
  \label{app_tab:da3_depth}
  \setlength{\tabcolsep}{7.5pt}
  \scriptsize
  \begin{tabular}{p{15pt}p{10pt}l rrrr rrrr}
  
    \toprule
    & & & \multicolumn{4}{c}{High overlap} & \multicolumn{4}{c}{Low overlap} \\
    \cmidrule(lr){4-7} \cmidrule(lr){8-11}
    & N & Method & \multicolumn{1}{c}{RMSE\,$\downarrow$} & \multicolumn{1}{c}{ARel\,$\downarrow$} & \multicolumn{1}{c}{$\delta$\,$\uparrow$} & \multicolumn{1}{c}{FID\,$\downarrow$}
            & \multicolumn{1}{c}{RMSE\,$\downarrow$} & \multicolumn{1}{c}{ARel\,$\downarrow$} & \multicolumn{1}{c}{$\delta$\,$\uparrow$} & \multicolumn{1}{c}{FID\,$\downarrow$} \\
    \midrule
    \multirow{8}{*}{\rotatebox[origin=c]{90}{one-sided}} & \multirow{2}{*}{$1$}
    & Ours-DA3  & 0.255 & 0.134 & 0.861 & 46.10 & 0.317 & 0.150 & 0.822 & 54.75 \\
    & & Ours-VGGT      & 0.232 & 0.124 & 0.874 & 45.98 & 0.319 & 0.149 & 0.846 & 53.78 \\
    \cmidrule(lr){2-11}
    & \multirow{2}{*}{$2$}
    & Ours-DA3  & 0.236 & 0.116 & 0.895 & 39.70 & 0.319 & 0.150 & 0.828 & 51.49 \\
    & & Ours-VGGT      & 0.209 & 0.102 & 0.909 & 39.63 & 0.293 & 0.124 & 0.876 & 49.09 \\
    \cmidrule(lr){2-11}
    & \multirow{2}{*}{$3$}
    & Ours-DA3  & 0.237 & 0.117 & 0.897 & 33.43 & 0.309 & 0.142 & 0.842 & 48.15 \\
    & & Ours-VGGT      & 0.213 & 0.101 & 0.913 & 32.70 & 0.290 & 0.120 & 0.875 & 46.48 \\
    \cmidrule(lr){2-11}
    & \multirow{2}{*}{$4$}
    & Ours-DA3 & 0.237 & 0.116 & 0.897 & 25.28 & 0.307 & 0.138 & 0.852 & 45.13 \\
    & & Ours-VGGT      & 0.218 & 0.103 & 0.905 & 24.80 & 0.287 & 0.120 & 0.871 & 43.95 \\
    \midrule
    \multirow{4}{*}{\rotatebox[origin=c]{90}{two-sided}} & \multirow{2}{*}{$2$}
    & Ours-DA3 & 0.236 & 0.105 & 0.898 & 24.32 & 0.307 & 0.147 & 0.828 & 42.47 \\
    & & Ours-VGGT      & 0.220 & 0.094 & 0.907 & 23.64 & 0.276 & 0.128 & 0.859 & 41.05 \\
    \cmidrule(lr){2-11}
    & \multirow{2}{*}{$4$}
    & Ours-DA3  & 0.218 & 0.092 & 0.919 & 13.13 & 0.247 & 0.114 & 0.881 & 27.09 \\
    & & Ours-VGGT      & 0.212 & 0.086 & 0.920 & 13.03 & 0.222 & 0.099 & 0.888 & 26.45 \\
    \bottomrule
  \end{tabular}
\end{table}

\subsection{ODE solver step count}
\label{supp_sub:odesteps}

We ablate the number of ODE integration steps at inference in~\cref{tab:ode_steps} on the one-sided $N{=}2$ ScanNet++ protocol for both
high- and low-overlap settings. Very low step counts reveal a clear difference between pixel-aligned metrics (observed regions) and distributional metrics (FID, unobserved regions). With only one to three steps, PSNR, SSIM, and depth errors are often strongest, but FID degrades substantially, particularly in the low-overlap regime. For example, RGB FID drops from $93.29$ at one step to $60.24$ at five steps and $58.35$ at ten
steps, while high-overlap FID similarly improves from $57.13$ to $42.89$ and $42.16$, respectively.

\begin{figure}
    \centering
    \includegraphics[width=\linewidth]{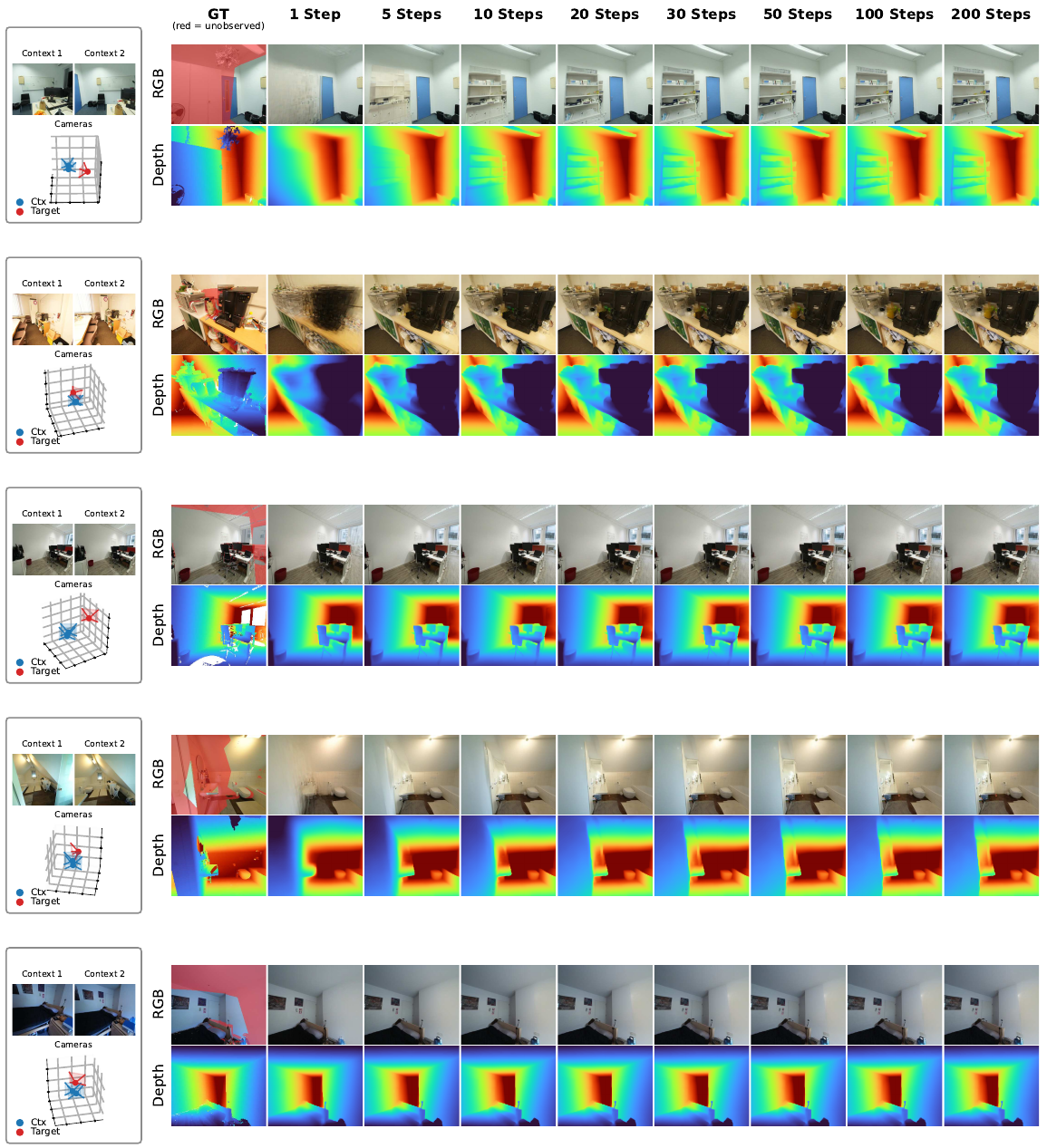} 

\caption{%
\textbf{Effect of ODE solver steps.}
Novel view generation on ScanNet++ using the one-sided protocol with two context views. Each example shows predicted RGB (top) and depth (bottom); red in the target marks regions unobserved by all context views. Across each row, the initial noise, context views, and target pose are fixed, and only the number of Euler integration steps changes. Very low step counts favor pixel-aligned reconstruction but produce visibly smoother generated regions, while additional steps recover finer appearance and object detail; beyond roughly $10$-$20$ steps, changes become increasingly subtle. Quantitative results are reported in \cref{tab:ode_steps}.%
}
    \label{fig:euler_steps_ablation}
\end{figure}

Beyond this low-step regime, performance is comparatively stable. From $5$ to $200$ steps, RGB PSNR varies by only $0.67$dB and $0.65$dB in the high-and low-overlap settings, respectively, while depth RMSE changes by $0.018$
and $0.042$. FID largely saturates after approximately $5$--$10$ steps, with
little benefit from substantially increasing the integration budget.
Pixel-aligned metrics gradually favor fewer steps, whereas qualitative
inspection shows that additional integration produces finer structures and
more detailed objects in regions not constrained by the context
(\cref{fig:euler_steps_ablation}). The effect is most apparent in generated regions, while visible-region reconstruction changes comparatively little.

We use $20$ steps for all main experiments as a practical compromise between
generation detail and computational efficiency. A $20$-step ODE solve takes
approximately $0.67$s per sample on an RTX A6000 ($\approx33$ms per ODE
step), plus $0.02$s and $0.04$s for depth and RGB decoding, respectively.
The strong performance obtained with fewer steps also allows inference to be
further accelerated when generation speed is prioritized.

\begin{table}[t]
\centering
\scriptsize
\setlength{\tabcolsep}{5pt}
\caption{%
\textbf{Effect of ODE solver steps.} Region-split RGB and depth metrics on the one-sided
ScanNet++ protocol with $N{=}2$ context views, for high and low overlap.
}
\begin{tabular}{@{}lr cccc cccc@{}}
\toprule
 & & \multicolumn{4}{c}{RGB} & \multicolumn{4}{c}{Depth} \\
\cmidrule(lr){3-6}\cmidrule(lr){7-10}
 & Steps & PSNR\,$\uparrow$& SSIM\,$\uparrow$ & LPIPS\,$\downarrow$ & FID\,$\downarrow$
      & RMSE\,$\downarrow$ & ARel\,$\downarrow$ & $\delta$\,$\uparrow$ & FID\,$\downarrow$ \\
\midrule
\multirow{10}{*}{\rotatebox[origin=c]{90}{High overlap}}
 & 1   & 20.27 & 0.707 & 0.086 & 57.13 & 0.188 & 0.093 & 0.917 & 44.84 \\
 & 2   & 20.09 & 0.708 & 0.083 & 47.87 & 0.192 & 0.093 & 0.919 & 42.08 \\
 & 3   & 19.88 & 0.703 & 0.085 & 44.02 & 0.194 & 0.094 & 0.919 & 40.92 \\
 & 5   & 19.62 & 0.696 & 0.088 & 42.89 & 0.198 & 0.096 & 0.916 & 39.98 \\
 & 10  & 19.35 & 0.687 & 0.091 & 42.16 & 0.204 & 0.099 & 0.913 & 39.55 \\
 & 20  & 19.13 & 0.681 & 0.096 & 42.16 & 0.207 & 0.102 & 0.914 & 39.66 \\
 & 30  & 19.08 & 0.678 & 0.095 & 42.81 & 0.212 & 0.104 & 0.907 & 39.46 \\
 & 50  & 19.02 & 0.675 & 0.096 & 43.17 & 0.214 & 0.105 & 0.904 & 39.52 \\
 & 100 & 18.97 & 0.673 & 0.097 & 43.44 & 0.215 & 0.106 & 0.903 & 39.67 \\
 & 200 & 18.95 & 0.672 & 0.098 & 43.66 & 0.216 & 0.107 & 0.902 & 39.75 \\
\midrule
\multirow{10}{*}{\rotatebox[origin=c]{90}{Low overlap}}
 & 1   & 18.91 & 0.656 & 0.087 & 93.29 & 0.262 & 0.112 & 0.883 & 59.56 \\
 & 2   & 18.62 & 0.654 & 0.086 & 73.69 & 0.261 & 0.112 & 0.887 & 55.31 \\
 & 3   & 18.39 & 0.647 & 0.088 & 64.99 & 0.264 & 0.113 & 0.888 & 53.08 \\
 & 5   & 18.09 & 0.638 & 0.090 & 60.24 & 0.271 & 0.116 & 0.889 & 50.90 \\
 & 10  & 17.81 & 0.630 & 0.093 & 58.35 & 0.282 & 0.120 & 0.884 & 49.66 \\
 & 20  & 17.76 & 0.630 & 0.095 & 58.86 & 0.283 & 0.121 & 0.882 & 49.80 \\
 & 30  & 17.58 & 0.622 & 0.096 & 57.66 & 0.302 & 0.128 & 0.873 & 49.30 \\
 & 50  & 17.51 & 0.619 & 0.097 & 58.18 & 0.306 & 0.130 & 0.873 & 49.45 \\
 & 100 & 17.46 & 0.617 & 0.098 & 58.40 & 0.311 & 0.133 & 0.871 & 49.60 \\
 & 200 & 17.44 & 0.616 & 0.098 & 58.75 & 0.313 & 0.134 & 0.870 & 49.62 \\
\bottomrule
\end{tabular}
\label{tab:ode_steps}
\end{table}

\subsection{Additional qualitative results}

\begin{figure}
    \centering
    \includegraphics[width=\linewidth]{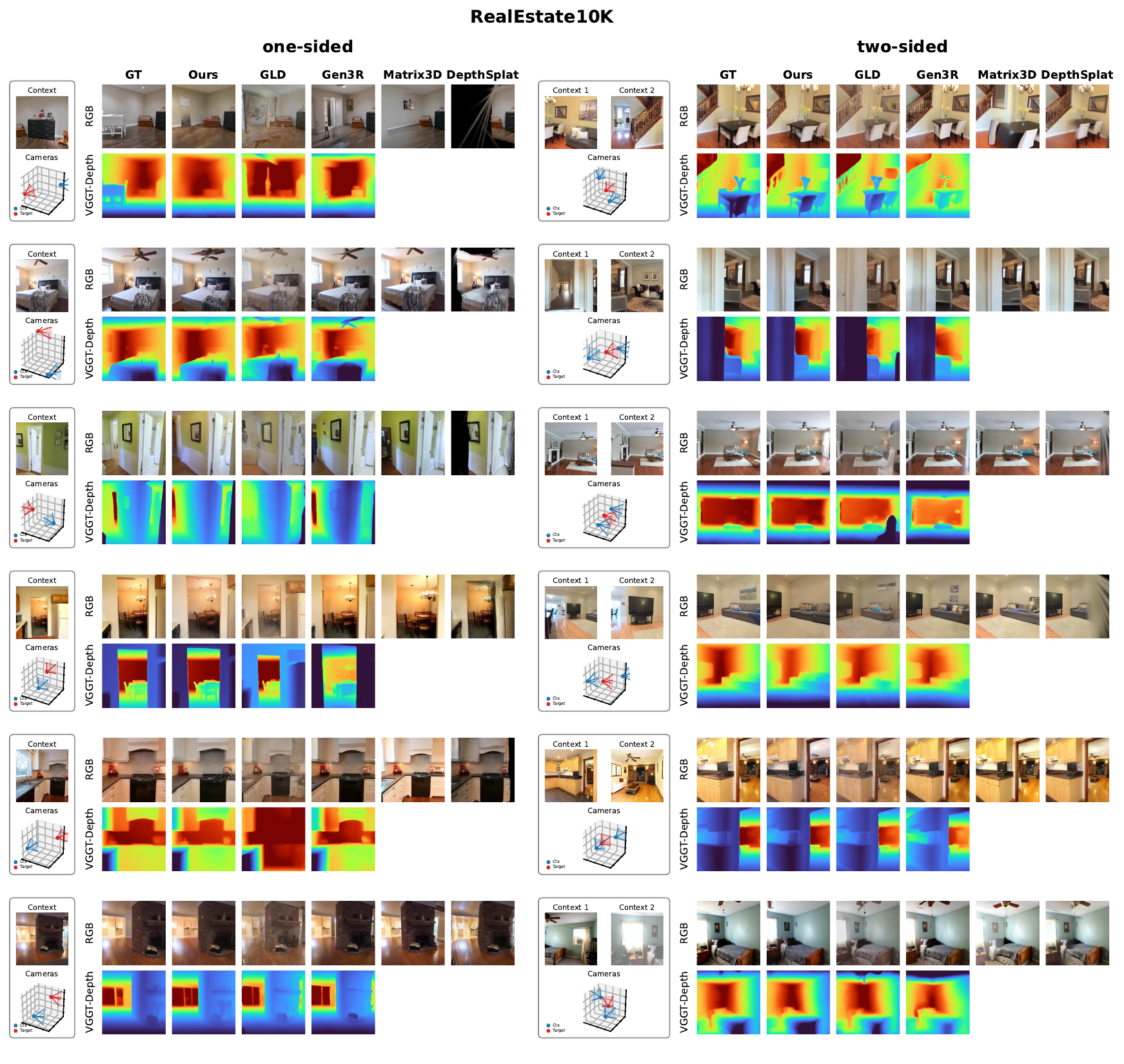} 
    \caption{\textbf{Additional qualitative results on RE10K.} One-sided (left) and two-sided (right) novel view generation. The context panel shows one (one-sided) or both (two-sided) context views and the camera layout (blue: context, red: target); columns show target RGB and VGGT pseudo-depth for the GT and all methods, the latter only for the VGGT-based methods.}
    \label{fig_app:nvs_re10k_qualitative}
\end{figure}

\begin{figure}
    \centering
    \includegraphics[width=\linewidth]{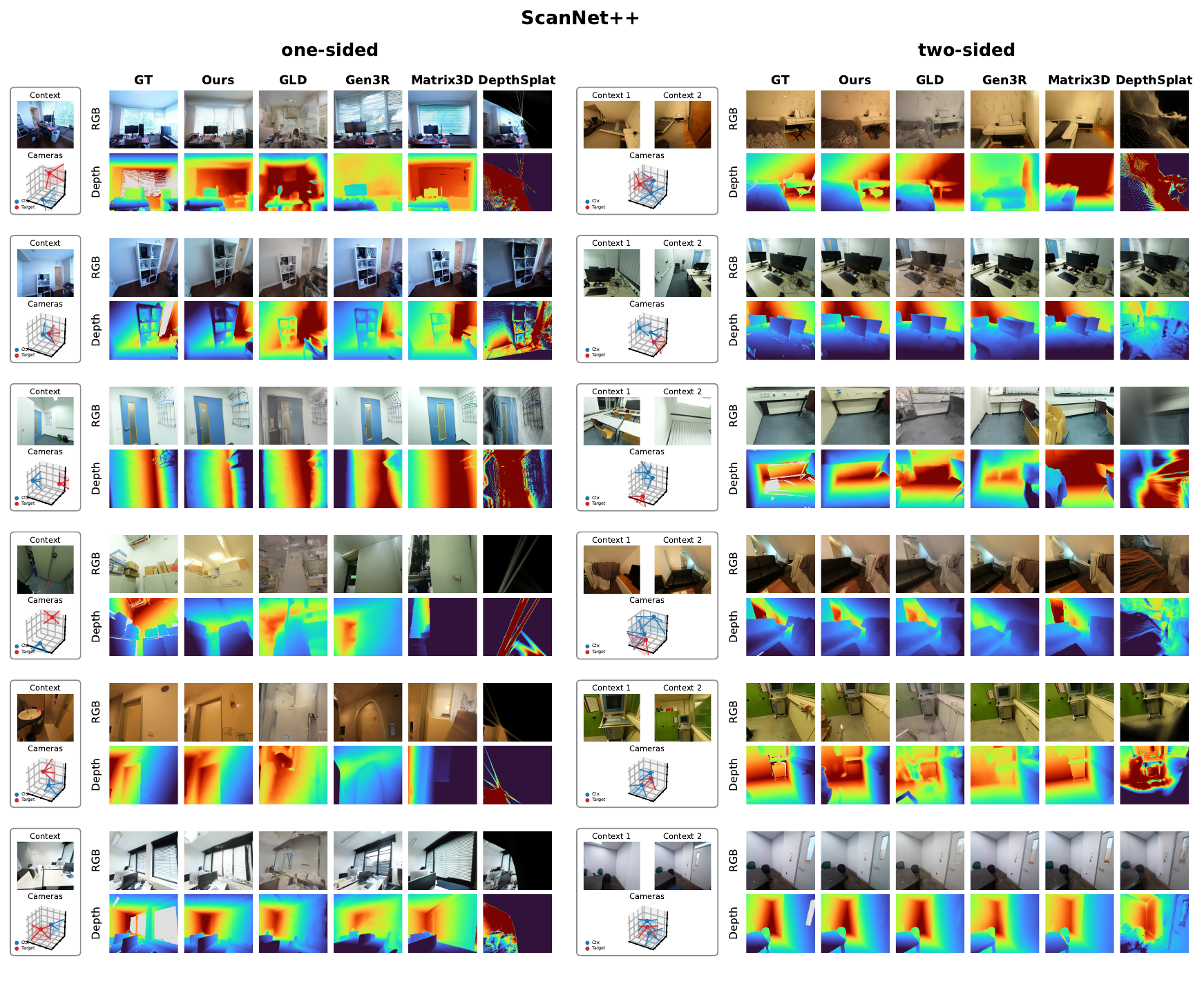} 
    \caption{\textbf{Additional qualitative results on ScanNet++.} One-sided (left) and two-sided (right) novel view generation. The context panel shows one (one-sided) or both (two-sided) context views and the camera layout (blue: context, red: target); columns show target RGB and depth for the GT and all methods.}
    \label{fig_app:nvs_scannet_qualitative}
\end{figure}

\begin{figure}[t]
    \centering
    \includegraphics[width=\linewidth]{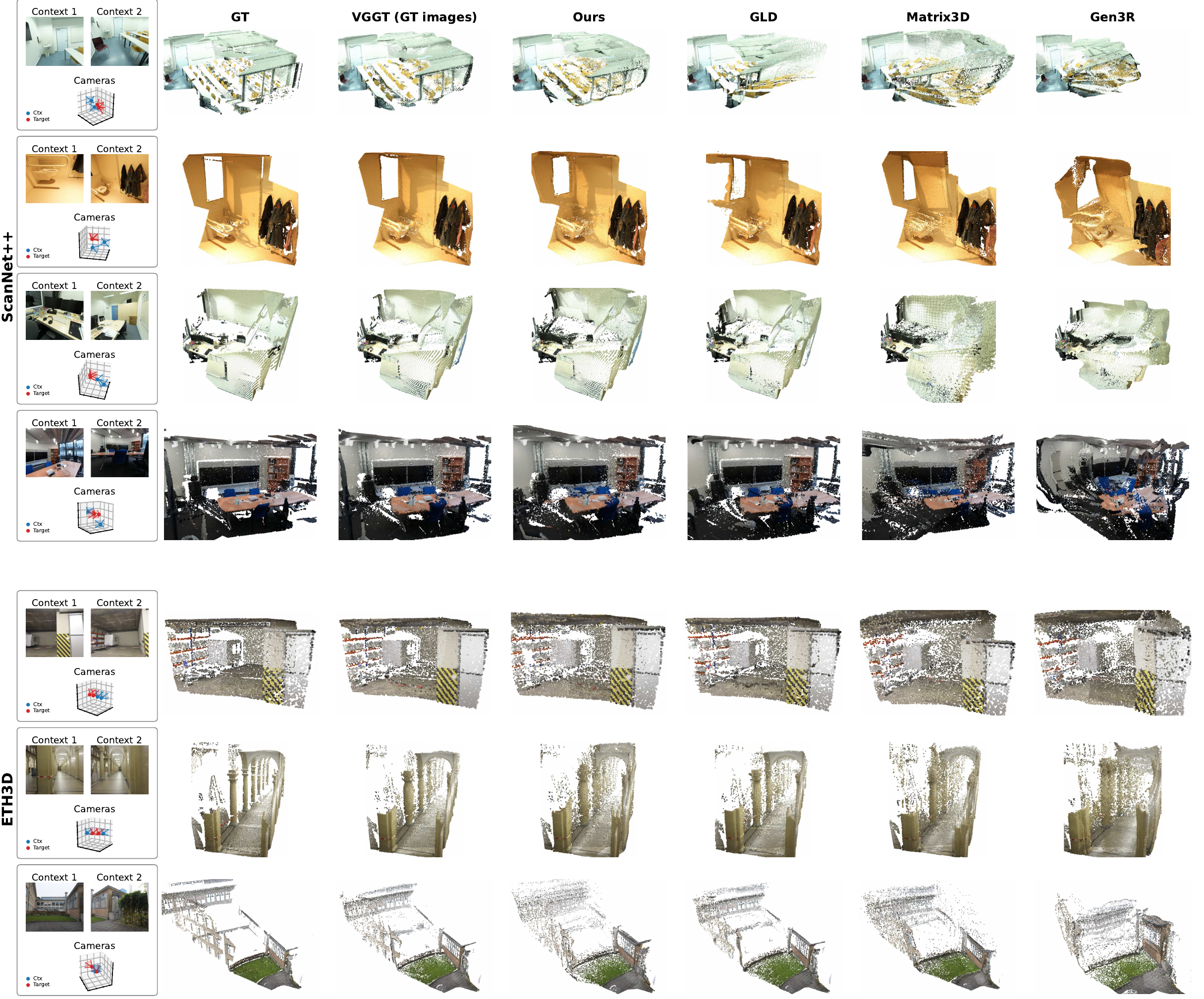}
    \caption{\textbf{Additional qualitative results for 3D scan generation on ScanNet++ (top) and ETH3D (bottom).} The context panel shows the two context views and the camera layout (blue: context, red: target); the columns show the point maps of the context and generated target views fused into a single scene, for the GT, for VGGT applied to the GT target images as an upper bound, and for all methods.}
    \label{fig_app:qualitative_3d}
\end{figure}

\Cref{fig_app:nvs_scannet_qualitative,fig_app:nvs_re10k_qualitative} show additional qualitative results for novel-view RGB and depth generation on ScanNet++ and RE10K in the one-sided and two-sided settings, and \cref{fig_app:qualitative_3d} shows further 3D scan generation results on ScanNet++ and ETH3D. The same observations as in \cref{fig:nvs_qualitative} hold across scenes: the reconstruction-based DepthSplat degrades sharply once the target leaves the support of the context views, GLD tends to blur appearance, Matrix3D and Gen3R render plausible images whose depth is coarse or inconsistent with the view, and our model keeps appearance and geometry consistent with each other and with the context, most visibly in the one-sided setting. For 3D scan generation (\cref{fig_app:qualitative_3d}), the picture matches \cref{tab:3d_eval}: the fused scans of our model and GLD stay close to the reference geometry on both datasets, including the much larger ETH3D scenes, whereas Matrix3D and Gen3R produce misaligned or duplicated structure that no longer lines up with the context point maps.

\subsection{Full tables of Euclidean vs. Riemannian flow matching results}
\label{app:eucl_vs_riemann}
\Cref{app_tab:eucl_riem} reports the full comparison between Euclidean and Riemannian flow matching across all context configurations for RGB (left) and depth (right). Respecting the latent geometry consistently improves performance: the Riemannian formulation outperforms the Euclidean formulation on every reported metric across all context counts and overlap regimes. For RGB, this yields substantial gains in both reconstruction quality (PSNR, SSIM, and LPIPS) and generation quality (FID). The same pattern holds for depth, with improvements in RMSE, ARel, $\delta$, and FID throughout.

The consistency of these gains across modalities, context regimes, and difficulty levels indicates that the benefit is not tied to a particular evaluation setting, but instead arises from performing flow matching on the latent manifold induced by the backbone representation rather than treating the latent as an unconstrained Euclidean space.

\begin{figure}
    \centering
    \includegraphics[width=\linewidth]{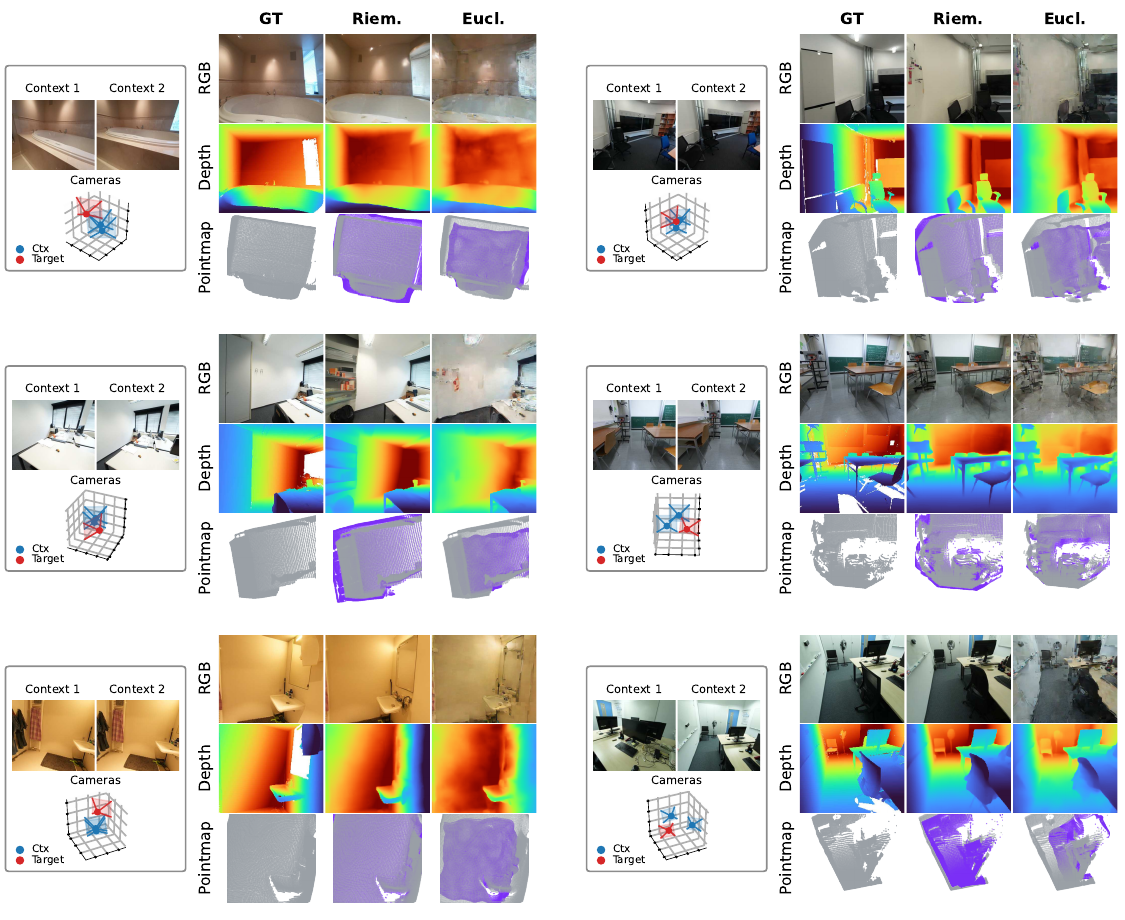} 

\caption{\textbf{Euclidean vs.\ Riemannian flow matching on ScanNet++} (two-sided, $N{=}4$). For each scene: the two context views and the camera layout (blue: context, red: target), and the target RGB, depth and fused point map (context points in gray, generated target points in color) for the GT, the Riemannian and the Euclidean model.}
    \label{fig_app:riem_vs_eucl_qual}
\end{figure}

\Cref{fig_app:riem_vs_eucl_qual} shows the two models side by side on six ScanNet++ scenes. The Euclidean samples decode to washed-out appearance and flattened depth, and their fused point maps bend away from the context geometry, whereas the Riemannian samples stay close to the ground truth in all three modalities.

\input{sec/tab_eucl_riem_full}

\paragraph{Off-subspace energy}
To quantify how far generated tokens leave the directions occupied by real decoder
inputs, let $P_b$ project onto the principal subspace holding $99\%$ of the centred
real-token variance of VGGT encoder block $b\in\{4,11,17,23\}$, with mean $\mu_b$.
The pooled off-subspace energy
\begin{equation}
  E_{\mathrm{off}}
  = \frac{\sum_{i,b}\bigl\lVert (I-P_b)\bigl(x_i^{(b)}-\mu_b\bigr)\bigr\rVert_2^2}
         {\sum_{i,b}\bigl\lVert x_i^{(b)}-\mu_b\bigr\rVert_2^2},
  \label{eq:off_subspace}
\end{equation}
where $i$ runs over the generated tokens, is the fraction of their variation that lies
outside the directions real tokens occupy. By construction it is $0.01$ for real tokens.

\Cref{app_tab:eucl_riem_endpoint} characterizes the generated tokens (two-sided,
$N{=}4$, low overlap configuration). The Euclidean endpoint leaves the data shell and
ends at $1.33R$. In addition, a third of its energy
($E_{\mathrm{off}}{=}0.33$) lies in directions real tokens do not occupy ($0.02$ for the
Riemannian model), and its cosine to the ground-truth token is $0.66$ against $0.85$.
After normalization, the token therefore sits on the shell, but at an untypical place,
and that is what the decoder sees. The Riemannian model never leaves the shell and its
direction is close to the real token's. Where this behavior comes from is analyzed in
\cref{app:trajectory_analysis}.

\begin{table}[t]
  \centering
  \caption{\textbf{Generated tokens under Euclidean and Riemannian flow matching.}
  Two-sided generation from $N{=}4$ context views on the low-overlap split of
  ScanNet++. Endpoint norm relative to the real-token radius $R$,
  cosine to the GT token, and off-subspace energy
  (\cref{eq:off_subspace}; real tokens $0.01$), the latter two in decoder-input
  coordinates. Best in \textbf{bold}.}
  \label{app_tab:eucl_riem_endpoint}
  \setlength{\tabcolsep}{6pt}
  \scriptsize
  \begin{tabular}{l c c c}
    \toprule
    Method & $\lVert x_1\rVert/R$ & $\cos(x_1,x_{\mathrm{gt}})\,\uparrow$ & $E_{\mathrm{off}}\,\downarrow$ \\
    \midrule
    Eucl.  & 1.332 & 0.657 & 0.334 \\
    Riem.  & 1.000 & {\bf 0.846} & {\bf 0.023} \\
    \bottomrule
  \end{tabular}
\end{table}

\subsection{Trajectory analysis: geometry and prediction target}
\label{app:trajectory_analysis}

The results in~\cref{app:eucl_vs_riemann} show that respecting the latent geometry substantially improves generation. We next disentangle the effects of geometry and prediction parameterization. Under Euclidean velocity prediction, the model must both reproduce the state-dependent term in the optimal velocity field and model the radial motion induced by straight-line transport. Endpoint prediction reparameterizes the same Euclidean flow so that the state term is constructed analytically by the sampler, removing the associated representational burden. However, the underlying path is unchanged and still contracts through the interior of the latent shell. Riemannian transport removes this radial component by construction. Finally, comparing velocity and endpoint prediction within the same Riemannian geometry isolates the remaining effect of the prediction objective itself.

\paragraph{Euclidean transport contains a radial component}
We write a latent token as $\token = \bigoplus_{i \in \{4, 11, 17, 23\}} h_i$,
where $\oplus$ is concatenation and $h_i \in \mathbb{R}^C$ is the output of VGGT's $i$th
encoder block.

The norm of ground-truth tokens of each encoder block, as well as the norm of the Gaussian source distribution tokens, is
strongly concentrated around an empirical radius $\lVert h_i \rVert_2 \approx R$. Since the product manifold of spheres is realized by concatenation of each sphere, we analyze one block in the following without loss of generality.

\begin{figure}
    \centering
    \includegraphics[width=0.9\linewidth]{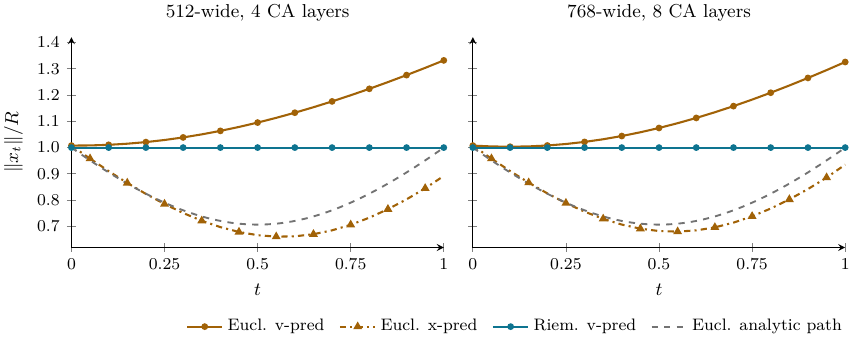}
    \caption{\textbf{Effect of geometry, prediction target, and model capacity on the sampling trajectory.} Normalized latent radius $\lVert x_t\rVert/R$ along the learned flow for the $(D,\#\mathrm{CA})=(512,4)$ and $(768,8)$ models. Riemannian flow remains on the manifold ($\lVert x_t\rVert/R=1$) by construction. Euclidean velocity prediction does not reproduce the radial component of the transport and drifts outward instead, with little change at larger capacity. Endpoint prediction removes this state-reconstruction burden and follows the analytic Euclidean chord substantially more closely; the larger model further reduces the discrepancy, but the trajectory still contracts through the interior of the latent shell.}
    \label{fig:trajectory}
\end{figure}

The Euclidean conditional path is the chord
\begin{equation}
    x_t=(1-t)x_0+t x_1, \qquad u_t=\dot{x}_t=x_1-x_0.
    \label{eq:euclidean_path}
\end{equation}
For $\lVert x_0\rVert_2\approx\lVert x_1\rVert_2\approx R$ and $\langle x_0,x_1\rangle\approx0$, as expected for high-dimensional source and target samples,
\begin{align}
    \frac{\lVert x_t\rVert_2^2}{R^2}
    &\approx
    (1-t)^2+t^2,\\
    \frac{\lVert x_t\rVert_2}{R}
    &\approx
    \sqrt{(1-t)^2+t^2}.
    \label{eq:chord}
\end{align}
Hence,
\begin{equation}
    \min_t\frac{\lVert x_t\rVert_2}{R}
    =
    \frac{1}{\sqrt{2}}
    \qquad\text{at}\qquad
    t=\frac12.
\end{equation}
This shows that although both endpoints lie approximately at the data shell, their Euclidean interpolation contracts to approximately $0.707R$ before expanding again (see the \textit{analytic path} curve in~\cref{fig:trajectory}).

The radial component can be made explicit by decomposing $u_t$ into radial and tangential components,
\begin{equation}
    u_t = u_t^{\mathrm{rad}} + u_t^{\mathrm{tan}}, \qquad u_t^{\mathrm{rad}}
    = \langle u_t,\hat{x}_t\rangle\hat{x}_t, \qquad \hat{x}_t=\frac{x_t}{\lVert x_t\rVert_2}.
\end{equation}
We define the radial fraction
\begin{equation}
    \rho(t) := \frac{\lVert u_t^{\mathrm{rad}}\rVert_2^2}
         {\lVert u_t\rVert_2^2}
    = \frac{\langle u_t,\hat{x}_t\rangle^2}
         {\lVert u_t\rVert_2^2}.
    \label{eq:rho}
\end{equation}
Intuitively, $\rho(t)$ measures the fraction of motion at time $t$ devoted to changing the radius, i.e.\ moving normal to the sphere, while $1-\rho(t)$ measures the fraction of motion along the sphere.

For the Euclidean conditional velocity $u_t=x_1-x_0$, the orthogonality approximation gives
\begin{equation}
    \rho_{\mathrm{Euc}}(t)
    \approx
    \frac{(2t-1)^2}
         {2\bigl((1-t)^2+t^2\bigr)},
    \label{eq:rho_euclidean}
\end{equation}
and therefore
\begin{equation}
    \rho_{\mathrm{Euc}}(0) = \rho_{\mathrm{Euc}}(1)=\frac12,
    \qquad
    \rho_{\mathrm{Euc}}\!\left(\frac12\right)=0.
\end{equation}
Thus, at either endpoint, approximately half of the squared Euclidean velocity is radial.
The Riemannian path instead follows a geodesic on the sphere. Its velocity satisfies
\begin{equation}
    u_t\in \T_{x_t}\mathcal{S}^{C-2},
    \qquad
    \langle u_t,x_t\rangle=0,
\end{equation}
and consequently
\begin{equation}
    \rho_{\mathrm{Riem}}(t)=0
    \qquad
    \forall t.
\end{equation}
The two geometries therefore define genuinely different transport problems:
Euclidean flow must model radial contraction and expansion, whereas
Riemannian flow uses all predicted motion in tangent directions.

\paragraph{Euclidean velocity prediction additionally requires state reconstruction}
For the linear path in \cref{eq:euclidean_path}, the MSE-optimal marginal velocity field is
\begin{equation}
    u^\star(x,t)
    =
    \mathbb{E}[x_1-x_0\mid x_t=x].
\end{equation}
Using $x_t=(1-t)x_0+t x_1$ gives
\begin{equation}
    x_0
    =
    \frac{x_t-tx_1}{1-t},
\end{equation}
and hence
\begin{equation}
    u^\star(x,t)
    =
    \frac{
        \mathbb{E}[x_1\mid x_t=x]-x
    }{1-t}.
    \label{eq:vstar}
\end{equation}

The target therefore, contains two different terms:
\begin{equation}
    u^\star(x,t)
    = \underbrace{
        \frac{\mathbb{E}[x_1\mid x_t=x]}{1-t}
    }_{\text{data-dependent}}
    - \underbrace{
        \frac{x}{1-t}
    }_{\text{state recon.}}.
    \label{eq:vstar_decomposition}
\end{equation}
The second term requires the network to reproduce the state, which is dominated by the isotropic source distribution sample near $t=0$.
The state term in \cref{eq:vstar_decomposition} is a $\damb$-dimensional isotropic draw, where $\damb=4C$ denotes the ambient token dimension, and a
network whose width drops to $D<\damb$ anywhere cannot reproduce it: its output is a function of a $D$-dimensional hidden state, so the set of outputs is a $D$-dimensional surface in $\mathbb{R}^{\damb}$ (Lipschitz maps do not increase dimension \cite[Cor.~2.4]{falconer2003fractal}), whatever the depth, attention or nonlinearities around the waist. With a linear output layer that surface is a subspace, the best the network can do is project onto it, and $1-D/\damb$ of the state's energy is lost ($91\%$ for $\damb=4C = 8192$, $D=768$). 

A nonlinear output layer can do better only by folding that surface, e.g., a curve snaking across a disc, so that it passes within $\delta$ of every point the noise can take, a $\damb$-dimensional volume of radius $\approx\sqrt{\damb}$. Covering that region takes $\sim(\sqrt{\damb}/\delta)^{\damb}$ balls of radius $\delta$, while the image of a bounded $D$-dimensional hidden state under a map of Lipschitz constant $L$ is covered by $\sim(L/\delta)^{D}$ of them \cite[Ch.~4]{vershynin2018high}, so matching the two requires an output sensitivity $L$ growing like $(\sqrt{\damb}/\delta)^{\damb/D-1}$, an exponent of $8192/768-1\approx9.7$ for our dimensions, far beyond the weight norms of trained networks, and consistent with the general result that networks narrower than their input are not universal approximators~\cite{lu2017expressive,hanin2017approximating,johnson2019deep,park2021minimum}.

Our measurements agree with the linear bound: with $\lVert x_0\rVert\approx R$, if the
uncancelled fraction $1-D/\damb$ of the noise energy were carried to the end orthogonally
to the data, the velocity model's endpoint would be
$\sqrt{R^2+(1-D/\damb)R^2}=\sqrt{2-D/\damb}\,R=1.38R$ ($D=768$) and $1.39R$ ($D=512$). We
measure $1.33R$ for both widths: the predicted magnitude, with no gain from the wider
model.

This provides a network width-dependent explanation for Euclidean velocity prediction. Its conditional transport requires radial motion, and its velocity parameterization asks the network itself to reconstruct the state needed to produce that motion. The Riemannian formulation instead removes the radial component from the prediction problem.

\paragraph{Endpoint prediction removes the state-reconstruction term, but not the Euclidean path}
Increasing the model width to the output dimension, can be circumvented by choosing a different prediction target. 
With endpoint prediction, the network estimates
\begin{equation}
    \hat{x}_1=f_\netpars(x_t,t,c,\pi),
\end{equation}
and the corresponding Euclidean velocity is assembled analytically as
\begin{equation}
    \hat{u}_t = \frac{\hat{x}_1-x_t}{1-t}.
    \label{eq:xpred_velocity}
\end{equation}
The full-dimensional $-x_t/(1-t)$ term is therefore computed directly by the sampler rather than reconstructed through the model. Consequently, the model width no longer limits this component.

This explains why Euclidean endpoint prediction performs substantially better than Euclidean velocity prediction. However, changing the prediction target does not change the geometry. If $\hat{x}_1=x_1$, integration of
\cref{eq:xpred_velocity} recovers the same straight conditional path
\begin{equation}
    x_t=(1-t)x_0+t x_1,
\end{equation}
and therefore the same contraction in \cref{eq:chord}. Endpoint prediction removes the need to \emph{learn} the radial state-reconstruction term, but the resulting transport still contains radial motion. This is consistent with the preference for clean-sample prediction reported for high-dimensional pixel space~\cite{li2025jit}, for frozen VGGT tokens~\cite{sun2026vggtworld} and for $\pi^3$ features~\cite{gao2026oneworld}: the parameterization removes the state-reconstruction term of the Euclidean velocity target, which recovers most of the gap, while the remaining difference to the Riemannian velocity flow at matched width (\cref{tab:width_scaling}) is attributable to the radial transport that the parameterization leaves in place.

\Cref{tab:width_scaling,fig:trajectory} further show how this effect interacts with model capacity. For Euclidean velocity prediction, increasing both width and the number of cross-attention layers yields only limited gains: the additional
depth cannot fully compensate for the narrow state bottleneck, since the
$8192$-dimensional input is still compressed to $D=768$. Euclidean endpoint prediction does not suffer from this state-reconstruction constraint and therefore benefits substantially more from the larger model. As a result, it
moves much closer to Riemannian velocity prediction at $D=768$ than at
$D=512$, although a gap remains. This is consistent with the view that
additional capacity can compensate for part of the Euclidean transport burden once the state-copy term is removed, whereas the Riemannian formulation eliminates the radial component by construction.

\begin{table}[t]
  \centering
  \caption{\textbf{Effect of flow geometry and prediction target across model size.} Two-sided generation from $N{=}4$ context views on the low-overlap split of ScanNet++. $D$ denotes the transformer width and \#CA the number of cross-attention layers. Best per model size in \textbf{bold}.}
  \label{tab:width_scaling}
  \setlength{\tabcolsep}{6pt}
  \scriptsize
  \begin{tabular}{p{18pt}p{16pt}l rrrr rrrr}
    \toprule
    & & & \multicolumn{4}{c}{RGB} & \multicolumn{4}{c}{Depth} \\
    \cmidrule(lr){4-7} \cmidrule(lr){8-11}
    $D$ & \#CA & Method & \multicolumn{1}{c}{PSNR\,$\uparrow$} & \multicolumn{1}{c}{SSIM\,$\uparrow$} & \multicolumn{1}{c}{LPIPS\,$\downarrow$} & \multicolumn{1}{c}{FID\,$\downarrow$}
           & \multicolumn{1}{c}{RMSE\,$\downarrow$} & \multicolumn{1}{c}{ARel\,$\downarrow$} & \multicolumn{1}{c}{$\delta$\,$\uparrow$} & \multicolumn{1}{c}{FID\,$\downarrow$} \\
    \midrule
    \multirow{3}{*}{$512$} & \multirow{3}{*}{$4$}
    & Eucl.-$v$    & 16.67 & 0.544 & 0.244 & 40.44 & 0.284 & 0.167 & 0.794 & 28.80 \\
    & & Eucl.-$x$    & 17.17 & 0.620 & 0.173 & 32.55 & 0.270 & 0.124 & 0.857 & 28.73 \\
    & & Riem.-$v$ & {\bf 17.80} & {\bf 0.639} & {\bf 0.156} & {\bf 31.49} & {\bf 0.242} & {\bf 0.108} & {\bf 0.891} & {\bf 27.74} \\
    \midrule
    \multirow{3}{*}{$768$} & \multirow{3}{*}{$8$}
    & Eucl.-$v$    & 16.98 & 0.553 & 0.232 & 39.00 & 0.292 & 0.178 & 0.775 & 28.68 \\
    & & Eucl.-$x$    & 18.11 & 0.647 & 0.143 & 29.84 & 0.239 & 0.107 & 0.897 & 26.27 \\
    & & Riem.-$v$ & {\bf 18.39} & {\bf 0.656} & {\bf 0.135} & {\bf 29.69} & {\bf 0.228} & {\bf 0.100} & {\bf 0.907} & {\bf 26.14} \\
    \bottomrule
  \end{tabular}
\end{table}

\paragraph{Why endpoint prediction also improves Riemannian flow}
We also train a Riemannian flow matching model with endpoint prediction, using the squared geodesic distance between the predicted and the true endpoint as the loss, with the prediction projected onto the manifold and converted to a velocity at sampling time.

The comparison between Riemannian velocity and endpoint prediction isolates a different effect: the geometry is identical in both cases and all velocities remain tangent, so the representability argument above does not apply. The
remaining difference arises from the prediction objective.

For the geodesic conditional path, the velocity target can be written as
\begin{equation}
    u_t
    =
    \frac{\Log_{x_t}(x_1)}{1-t}.
    \label{eq:riem_vtarget}
\end{equation}
Any tangent prediction $\hat{u}_t$ can be written in the same form through
the endpoint it implies, $\hat{x}_1=\Exp_{x_t}\!\big((1-t)\,\hat{u}_t\big)$,
so that
\begin{equation}
    \hat{u}_t
    =
    \frac{\Log_{x_t}(\hat{x}_1)}{1-t},
\end{equation}
and the velocity error is exactly
\begin{equation}
    \lVert\hat{u}_t-u_t\rVert_2^2
    =
    \frac{
        \left\lVert
        \Log_{x_t}(\hat{x}_1)
        -
        \Log_{x_t}(x_1)
        \right\rVert_2^2
    }{(1-t)^2}.
    \label{eq:velocity_weight_exact}
\end{equation}
For a small endpoint perturbation
$\delta=\hat{x}_1-x_1$, first-order expansion of the logarithmic map gives
\begin{equation}
    \Log_{x_t}(\hat{x}_1)
    -
    \Log_{x_t}(x_1)
    =
    J_{\Log_{x_t}}(x_1)\,\delta
    +
    O(\lVert\delta\rVert_2^2),
\end{equation}
and therefore
\begin{equation}
    \lVert\hat{u}_t-u_t\rVert_2^2
    \approx
    \frac{
        \left\lVert
        J_{\Log_{x_t}}(x_1)\,\delta
        \right\rVert_2^2
    }{(1-t)^2}.
    \label{eq:weight}
\end{equation}

Velocity regression is therefore an endpoint regression with an explicit
$(1-t)^{-2}$ weight, whereas the geodesic endpoint loss carries weight one. The
weight matters because of what the velocity target becomes late on the path. As
$t\rightarrow1$ the residual $x_1-x_t$ vanishes while the model's uncertainty about
$x_1$ does not, so the direction of that residual, which is the velocity target,
can no longer be resolved from the state: the relative velocity error
$\lVert v_w-u_t\rVert^2/\lVert u_t\rVert^2$ of the trained model rises from
$0.09$ at $t{=}0.2$ to $0.99$ at $t{=}0.99$, the value of an uninformative
prediction, whereas the geodesic loss of the endpoint model falls monotonically
and is flat for $t\ge0.8$. At the same time a late velocity error barely affects
the sample, since it enters only through the remaining path length
$\propto(1-t)$. Under the shifted time sampler the velocity objective nevertheless
spends $11\%$ of its expected loss on $t>0.9$ and $37\%$ on $t>0.5$, against
$1\%$ and $11\%$ for the endpoint objective: a substantial part of its gradient
goes where the target is unresolvable and the sample insensitive, a regime the
endpoint loss de-weights automatically. The effect on the final samples is modest
(\cref{tab:pred_target}), because the velocity objective still draws most of its
signal from the early path.

\begin{table}[t]
  \centering
  \caption{\textbf{Prediction target under Riemannian flow matching.} Two-sided generation from $N{=}4$ context views on the low-overlap split of ScanNet++. $v$ denotes velocity prediction, $x$ sample prediction. Best in \textbf{bold}.}
  \label{tab:pred_target}
  \setlength{\tabcolsep}{6pt}
  \scriptsize
  \begin{tabular}{l rrrr rrrr}
    \toprule
    & \multicolumn{4}{c}{RGB} & \multicolumn{4}{c}{Depth} \\
    \cmidrule(lr){2-5} \cmidrule(lr){6-9}
    Method & \multicolumn{1}{c}{PSNR\,$\uparrow$} & \multicolumn{1}{c}{SSIM\,$\uparrow$} & \multicolumn{1}{c}{LPIPS\,$\downarrow$} & \multicolumn{1}{c}{FID\,$\downarrow$}
           & \multicolumn{1}{c}{RMSE\,$\downarrow$} & \multicolumn{1}{c}{ARel\,$\downarrow$} & \multicolumn{1}{c}{$\delta$\,$\uparrow$} & \multicolumn{1}{c}{FID\,$\downarrow$} \\
    \midrule
    Riem.-$v$ & 18.39 & 0.656 & 0.135 & 29.69 & 0.228 & 0.100 & {\bf 0.907} & 26.14 \\
    Riem.-$x$ & {\bf 18.56} & {\bf 0.662} & {\bf 0.130} & {\bf 28.61} & {\bf 0.227} & {\bf 0.099} & 0.905 & {\bf 25.92} \\
    \bottomrule
  \end{tabular}
\end{table}

%% file: imgs/protocols.tex
\begin{figure}[!t]
\centering
\resizebox{\textwidth}{!}{
\begin{tikzpicture}[
box/.style={draw, thick, rounded corners=2pt, minimum width=1.4cm, minimum height=0.8cm, align=center, font=\footnotesize\sffamily},
ourRGB/.style={box, fill=cyan!15, draw=cyan!60!black},
target/.style={box, fill=orange!20, draw=orange!80!black, thick},
ourArr/.style={->, >=Stealth, thick, cyan!70!black},
jumpArr/.style={->, >=Stealth, thick, cyan!70!black, dashed}
]

\node[font=\large\bfseries\sffamily] at (4, 1.5) {One-sided setup};
\node[font=\large\bfseries\sffamily] at (15, 1.5) {Two-sided setup};

\node[font=\normalsize\bfseries\sffamily, anchor=east] at (-1.5, 0.0) {$N=1$};
\node[font=\normalsize\bfseries\sffamily, anchor=east] at (-1.5, -1.5) {$N=2$};
\node[font=\normalsize\bfseries\sffamily, anchor=east] at (-1.5, -3.0) {$N=3$};
\node[font=\normalsize\bfseries\sffamily, anchor=east] at (-1.5, -4.5) {$N=4$};

\draw[gray!30, thick] (9.5, 2.0) -- (9.5, -5.5);

\node[ourRGB] (o1_e_1) at (0, 0.0) {Context\\RGB};
\node[target] (o1_e_tgt) at (8, 0.0) {Target};
\draw[jumpArr] (o1_e_1) -- (o1_e_tgt);

\node[ourRGB] (o2_e_1) at (0, -1.5) {Context\\RGB};
\node[ourRGB] (o2_e_2) at (2, -1.5) {Context\\RGB};
\node[target] (o2_e_tgt) at (8, -1.5) {Target};
\draw[ourArr] (o2_e_1) -- (o2_e_2);
\draw[jumpArr] (o2_e_2) -- (o2_e_tgt);

\node[ourRGB] (o3_e_1) at (0, -3.0) {Context\\RGB};
\node[ourRGB] (o3_e_2) at (2, -3.0) {Context\\RGB};
\node[ourRGB] (o3_e_3) at (4, -3.0) {Context\\RGB};
\node[target] (o3_e_tgt) at (8, -3.0) {Target};
\draw[ourArr] (o3_e_1) -- (o3_e_2);
\draw[ourArr] (o3_e_2) -- (o3_e_3);
\draw[jumpArr] (o3_e_3) -- (o3_e_tgt);

\node[ourRGB] (o4_e_1) at (0, -4.5) {Context\\RGB};
\node[ourRGB] (o4_e_2) at (2, -4.5) {Context\\RGB};
\node[ourRGB] (o4_e_3) at (4, -4.5) {Context\\RGB};
\node[ourRGB] (o4_e_4) at (6, -4.5) {Context\\RGB};
\node[target] (o4_e_tgt) at (8, -4.5) {Target};
\draw[ourArr] (o4_e_1) -- (o4_e_2);
\draw[ourArr] (o4_e_2) -- (o4_e_3);
\draw[ourArr] (o4_e_3) -- (o4_e_4);
\draw[jumpArr] (o4_e_4) -- (o4_e_tgt);

\node[ourRGB] (o2_i_1) at (11, -1.5) {Context\\RGB};
\node[target] (o2_i_tgt) at (15, -1.5) {Target};
\node[ourRGB] (o2_i_2) at (19, -1.5) {Context\\RGB};
\draw[jumpArr] (o2_i_1) -- (o2_i_tgt);
\draw[jumpArr] (o2_i_2) -- (o2_i_tgt);

\node[ourRGB] (o4_i_1) at (11, -4.5) {Context\\RGB};
\node[ourRGB] (o4_i_2) at (13, -4.5) {Context\\RGB};
\node[target] (o4_i_tgt) at (15, -4.5) {Target};
\node[ourRGB] (o4_i_3) at (17, -4.5) {Context\\RGB};
\node[ourRGB] (o4_i_4) at (19, -4.5) {Context\\RGB};
\draw[ourArr] (o4_i_1) -- (o4_i_2);
\draw[jumpArr] (o4_i_2) -- (o4_i_tgt);
\draw[ourArr] (o4_i_4) -- (o4_i_3);
\draw[jumpArr] (o4_i_3) -- (o4_i_tgt);

\end{tikzpicture}
}
\caption{\textbf{Evaluation setup.} We use two setups, \emph{one-sided} and \emph{two-sided} setups. 
In \emph{one-sided}, we provide the model a set of consecutive frames within a trajectory without pose and task the model to predict a target view defined by the relative pose to the first frame.
We provide experiments for a variable context length, $N \in \{1, 2, 3, 4\}$.
In \emph{two-sided}, we provide unposed frames from the beginning and end of a trajectory and request the model to predict an intermediate frame of such a trajectory.
We evaluate model with two context lengths, $N \in \{2, 4\}$.}
\label{fig:sequence_layout}
\end{figure}

%% file: sec/tab_eucl_riem_full.tex
\begin{table}[t]
  \providecommand{\metrichead}[1]{{\fontsize{6pt}{7pt}\selectfont #1}}
  \centering
  \caption{\textbf{Ablation on flow geometry on ScanNet++, RGB (left) and depth (right).} Euclidean and Riemannian flow matching under a matched architecture and training setup. Best in \textbf{bold}.}
  \label{app_tab:eucl_riem}
  \setlength{\tabcolsep}{1.4pt}
  \setlength{\aboverulesep}{0pt}\setlength{\belowrulesep}{0pt}
  \scriptsize
  \begin{tabular}[t]{p{9pt}p{6pt}l@{\hspace{5pt}}cccc cccc}
    \toprule
    & & & \multicolumn{8}{c}{RGB} \\
    \cmidrule(lr){4-11}
    & & & \multicolumn{4}{c}{\rule{0pt}{2.3ex}High overlap} & \multicolumn{4}{c}{Low overlap} \\
    \cmidrule(lr){4-7} \cmidrule(lr){8-11}
    & N & \rule{0pt}{2.3ex}Method & \metrichead{PSNR$\uparrow$} & \metrichead{SSIM$\uparrow$} & \metrichead{LPIPS$\downarrow$} & \metrichead{FID$\downarrow$} & \metrichead{PSNR$\uparrow$} & \metrichead{SSIM$\uparrow$} & \metrichead{LPIPS$\downarrow$} & \metrichead{FID$\downarrow$} \\
    \midrule
    \multirow{8}{*}{\rotatebox[origin=c]{90}{one-sided}} & \multirow{2}{*}{$1$}
    & \rule{0pt}{2.3ex}Eucl.     & 17.25 & 0.566 & 0.142 & 78.33 & 16.30 & 0.522 & 0.120 & 104.17 \\
    & & Riem.     & \textbf{18.99} & \textbf{0.677} & \textbf{0.085} & \textbf{49.73} & \textbf{17.61} & \textbf{0.620} & \textbf{0.086} & \textbf{63.86} \\[1.2pt]
    \cmidrule(lr){2-11}
    & \multirow{2}{*}{$2$}
    & \rule{0pt}{2.3ex}Eucl.     & 17.27 & 0.565 & 0.167 & 63.53 & 16.43 & 0.531 & 0.136 & 93.09 \\
    & & Riem.     & \textbf{19.13} & \textbf{0.681} & \textbf{0.096} & \textbf{42.16} & \textbf{17.76} & \textbf{0.630} & \textbf{0.095} & \textbf{58.86} \\[1.2pt]
    \cmidrule(lr){2-11}
    & \multirow{2}{*}{$3$}
    & \rule{0pt}{2.3ex}Eucl.     & 17.42 & 0.569 & 0.190 & 50.88 & 16.49 & 0.534 & 0.150 & 86.61 \\
    & & Riem.     & \textbf{19.35} & \textbf{0.685} & \textbf{0.105} & \textbf{35.30} & \textbf{17.78} & \textbf{0.632} & \textbf{0.102} & \textbf{55.10} \\[1.2pt]
    \cmidrule(lr){2-11}
    & \multirow{2}{*}{$4$}
    & \rule{0pt}{2.3ex}Eucl.     & 17.57 & 0.572 & 0.217 & 36.16 & 16.48 & 0.535 & 0.162 & 78.37 \\
    & & Riem.     & \textbf{19.59} & \textbf{0.691} & \textbf{0.113} & \textbf{25.67} & \textbf{17.73} & \textbf{0.632} & \textbf{0.110} & \textbf{51.86} \\[1.2pt]
    \midrule
    \multirow{4}{*}{\rotatebox[origin=c]{90}{two-sided}} & \multirow{2}{*}{$2$}
    & \rule{0pt}{2.3ex}Eucl.     & 17.40 & 0.564 & 0.217 & 36.53 & 16.38 & 0.537 & 0.180 & 71.84 \\
    & & Riem.     & \textbf{19.58} & \textbf{0.687} & \textbf{0.112} & \textbf{26.10} & \textbf{17.43} & \textbf{0.631} & \textbf{0.121} & \textbf{50.67} \\[1.2pt]
    \cmidrule(lr){2-11}
    & \multirow{2}{*}{$4$}
    & \rule{0pt}{2.3ex}Eucl.     & 17.77 & 0.578 & 0.247 & 17.42 & 16.98 & 0.553 & 0.232 & 39.00 \\
    & & Riem.     & \textbf{20.18} & \textbf{0.703} & \textbf{0.116} & \textbf{12.89} & \textbf{18.39} & \textbf{0.656} & \textbf{0.135} & \textbf{29.69} \\[1.2pt]
    \bottomrule
  \end{tabular}%
  \hspace{6pt}%
  \begin{tabular}[t]{cccc cccc}
    \toprule
    \multicolumn{8}{c}{Depth} \\
    \cmidrule(lr){1-8}
    \multicolumn{4}{c}{\rule{0pt}{2.3ex}High overlap} & \multicolumn{4}{c}{Low overlap} \\
    \cmidrule(lr){1-4} \cmidrule(lr){5-8}
    \rule{0pt}{2.3ex}\metrichead{RMSE$\downarrow$} & \metrichead{ARel$\downarrow$} & \metrichead{$\delta$$\uparrow$} & \metrichead{FID$\downarrow$} & \metrichead{RMSE$\downarrow$} & \metrichead{ARel$\downarrow$} & \metrichead{$\delta$$\uparrow$} & \metrichead{FID$\downarrow$} \\
    \midrule
    \rule{0pt}{2.3ex}0.271 & 0.169 & 0.820 & 50.08 & 0.358 & 0.187 & 0.761 & 61.67 \\
    \textbf{0.223} & \textbf{0.118} & \textbf{0.891} & \textbf{45.99} & \textbf{0.299} & \textbf{0.138} & \textbf{0.852} & \textbf{53.55} \\[1.2pt]
    \cmidrule(lr){1-8}
    \rule{0pt}{2.3ex}0.265 & 0.161 & 0.825 & 42.61 & 0.345 & 0.176 & 0.781 & 56.45 \\
    \textbf{0.207} & \textbf{0.102} & \textbf{0.914} & \textbf{39.66} & \textbf{0.283} & \textbf{0.121} & \textbf{0.882} & \textbf{49.80} \\[1.2pt]
    \cmidrule(lr){1-8}
    \rule{0pt}{2.3ex}0.275 & 0.164 & 0.823 & 35.61 & 0.346 & 0.177 & 0.776 & 52.15 \\
    \textbf{0.211} & \textbf{0.100} & \textbf{0.915} & \textbf{32.89} & \textbf{0.282} & \textbf{0.120} & \textbf{0.876} & \textbf{46.58} \\[1.2pt]
    \cmidrule(lr){1-8}
    \rule{0pt}{2.3ex}0.288 & 0.171 & 0.813 & 26.07 & 0.354 & 0.182 & 0.764 & 48.69 \\
    \textbf{0.217} & \textbf{0.102} & \textbf{0.907} & \textbf{24.73} & \textbf{0.280} & \textbf{0.118} & \textbf{0.876} & \textbf{44.04} \\[1.2pt]
    \midrule
    \rule{0pt}{2.3ex}0.276 & 0.156 & 0.828 & 25.62 & 0.328 & 0.193 & 0.751 & 44.84 \\
    \textbf{0.220} & \textbf{0.094} & \textbf{0.908} & \textbf{23.64} & \textbf{0.276} & \textbf{0.128} & \textbf{0.856} & \textbf{41.41} \\[1.2pt]
    \cmidrule(lr){1-8}
    \rule{0pt}{2.3ex}0.272 & 0.152 & 0.832 & 13.69 & 0.292 & 0.178 & 0.775 & 28.68 \\
    \textbf{0.211} & \textbf{0.085} & \textbf{0.921} & \textbf{13.23} & \textbf{0.228} & \textbf{0.100} & \textbf{0.907} & \textbf{26.14} \\[1.2pt]
    \bottomrule
  \end{tabular}
\end{table}